\documentclass{article}

\usepackage{pifont}

\usepackage[utf8]{inputenc} %
\usepackage[T1]{fontenc}    %
\usepackage{hyperref}       %
\usepackage{url}            %
\usepackage{booktabs}       %
\usepackage{amsfonts}       %
\usepackage{nicefrac}       %
\usepackage{microtype}      %
\usepackage{xcolor}         %

\usepackage[accepted]{icml2025}

\newcommand{\email}[1]{\href{mailto:#1}{#1}}

\usepackage{amsmath,amsfonts,bm}

\def\eqref#1{equation~\ref{#1}}

\def\1{\bm{1}}

\DeclareMathAlphabet{\mathsfit}{\encodingdefault}{\sfdefault}{m}{sl}
\SetMathAlphabet{\mathsfit}{bold}{\encodingdefault}{\sfdefault}{bx}{n}

\def\gL{{\mathcal{L}}}

\newcommand{\E}{\mathbb{E}}

\DeclareMathOperator*{\argmin}{arg\,min}

\usepackage[utf8]{inputenc} %
\usepackage{lipsum} %

\usepackage[T1]{fontenc}

\usepackage{etoolbox}

\usepackage{etoolbox}

\newbool{includeappendix}
\setbool{includeappendix}{true} %
\IfFileExists{headers/config/noappendix.config}{
	\setbool{includeappendix}{false}
}{}

\newif\ifincludeappendixx
\ifbool{includeappendix}{
	\includeappendixxtrue
}{
	\includeappendixxfalse
}

\usepackage{xr} %
\usepackage{filecontents}

\ifbool{includeappendix}{}{
	\input{appendix-labels-loader}

	\externaldocument{appendix-labels}
}

\ifdefined\isoverfull
\else
\fi

\usepackage{subcaption}

\usepackage{xcolor} %

\definecolor{my-full-blue}{HTML}{1F77B4}

\definecolor{my-full-orange}{HTML}{FF7F0E}

\definecolor{my-full-green}{HTML}{2CA02C}

\definecolor{my-full-red}{HTML}{d62728}

\definecolor{my-full-purple}{HTML}{9467bd}

\definecolor{my-full-brown}{HTML}{8c564b}

\definecolor{my-full-pink}{HTML}{e377c2}

\definecolor{my-full-gray}{HTML}{7f7f7f}

\definecolor{my-full-olive}{HTML}{bcbd22}

\definecolor{my-full-cyan}{HTML}{17becf}

\definecolor{c1}{RGB}{86, 100, 26}
\definecolor{c2}{RGB}{192, 175, 251}
\definecolor{c3}{RGB}{230, 161, 118}
\definecolor{c4}{RGB}{0, 103, 138}
\definecolor{c5}{RGB}{152, 68, 100}
\definecolor{c6}{RGB}{94, 204, 171}
\definecolor{c7}{RGB}{205, 205, 205}

\definecolor{csabr}{RGB}{152, 68, 100}
\definecolor{cibp}{RGB}{94, 204, 171}
\definecolor{ctaps}{RGB}{0, 103, 138}

\definecolor{cm1}{HTML}{1f77b4}
\definecolor{cm2}{HTML}{ff7f0e}
\definecolor{cm3}{HTML}{2ca02c}
\definecolor{cm4}{HTML}{d62728}
\definecolor{cm5}{HTML}{9467bd}
\definecolor{cm6}{HTML}{8c564b}
\definecolor{cm7}{HTML}{e377c2}
\definecolor{cm8}{HTML}{7f7f7f}
\definecolor{cm9}{HTML}{bcbd22}
\definecolor{cm10}{HTML}{17becf}

\colorlet{my-blue}{my-full-blue!30}
\colorlet{my-orange}{my-full-orange!30}
\colorlet{my-green}{my-full-green!30}
\colorlet{my-red}{my-full-red!30}
\colorlet{my-purple}{my-full-purple!30}
\colorlet{my-brown}{my-full-brown!30}
\colorlet{my-pink}{my-full-pink!30}
\colorlet{my-gray}{my-full-gray!30}
\colorlet{my-olive}{my-full-olive!30}
\colorlet{my-cyan}{my-full-cyan!30}

\usepackage{listings}

\usepackage{textcomp}

\usepackage{xcolor}

\usepackage[scaled=0.8]{beramono}

\definecolor{ckeyword}{HTML}{7F0055}
\definecolor{ccomment}{HTML}{3F7F5F}
\definecolor{cstring}{HTML}{2A0099}

\lstdefinestyle{numbers}{
	numbers=left,
	framexleftmargin=20pt,
	numberstyle=\tiny,
	firstnumber=auto,
	numbersep=1em,
	xleftmargin=2em
}

\lstdefinestyle{layout}{
	frame=none,
	captionpos=b,
}

\lstdefinestyle{comment-style}{
	morecomment=[l]//,
	morecomment=[s]{/*}{*/},
	commentstyle={\color{ccomment}\itshape},
}

\lstdefinestyle{string-style}{
	morestring=[b]",%
	morestring=[b]',%
	stringstyle={\color{cstring}},
	showstringspaces=false,%
}

\lstdefinestyle{keyword-style}{
	keywordstyle={\ttfamily\bfseries},
	morekeywords={
		function,
		constructor,
		int,
		bool,
		return,
		returns,
		uint
	},
	morekeywords = [2]{},
	keywordstyle = [2]{\text},
	sensitive=true,
}

\lstdefinestyle{input-encoding}{
	inputencoding=utf8,
	extendedchars=true,
	literate=
	{ℝ}{$\reals$}1%
	{→}{$\rightarrow$}1%
	{α}{$\alpha$}1%
	{β}{$\beta$}1%
	{λ}{$\lambda$}1%
	{θ}{$\theta$}1%
	{ϕ}{$\phi$}1%
}

\lstdefinestyle{escaping}{
	moredelim={**[is][\color{blue}]{\%}{\%}},
	escapechar=|,
	mathescape=true
}

\lstdefinestyle{default-style}{
	basicstyle=\fontencoding{T1}\ttfamily\footnotesize,
	style=numbers,
	style=layout,
	style=comment-style,
	style=string-style,
	style=keyword-style,
	style=input-encoding,
	style=escaping,
	tabsize=2,
	upquote=true
}

\lstdefinelanguage{BASIC}{
	language=C++,
	style=default-style
}[keywords,comments,strings]%

\usepackage[capitalize,noabbrev]{cleveref}

\crefname{listing}{Lst.}{listings}
\crefname{line}{Lin.}{Lin.}
\crefname{appendix}{App.}{App.}
\crefname{lemma}{Lemma}{Lemmas}
\Crefname{lemma}{Lemma}{Lemmas}
\crefname{thm}{Theorem}{Theorems}
\Crefname{thm}{Theorem}{Theorems}

\newcommand{\app}[1]{%
	\ifbool{includeappendix}{\cref{#1}}{the appendix}%
}
\newcommand{\App}[1]{%
	\ifbool{includeappendix}{\cref{#1}}{The appendix}%
}

\usepackage{tikz}

\usetikzlibrary{calc,decorations,decorations.pathmorphing}
\usetikzlibrary{positioning,fit,arrows}
\usetikzlibrary{decorations.markings}
\usetikzlibrary{shapes,shapes.geometric}
\usetikzlibrary{shadows,patterns,snakes}
\usetikzlibrary{backgrounds,decorations.pathreplacing,automata}
\usetikzlibrary{intersections}
\usetikzlibrary{angles,quotes}
\usetikzlibrary{plotmarks}
\usetikzlibrary{patterns}
\usetikzlibrary{arrows.meta}
\usetikzlibrary{shapes.misc}
\usetikzlibrary{chains}
\usetikzlibrary{calc}

\usepackage[utf8]{inputenc} %
\usepackage[T1]{fontenc}    %
\usepackage{url}            %
\usepackage{booktabs}       %
\usepackage{amsfonts}       %
\usepackage{nicefrac}       %
\usepackage{microtype}      %

\usepackage{graphicx}
\usepackage{adjustbox}
\usepackage{threeparttable}
\usepackage{thmtools, nameref}
\usepackage{multirow}
\usepackage{amssymb}
\usepackage{mathtools}
\usepackage{amsthm}
\usepackage[abbreviations]{foreign}
\usepackage{enumitem}
\usepackage{makecell}
\usepackage{wrapfig}
\usepackage{subcaption}

\crefname{lemma}{Lemma}{lemmas}
\Crefname{lemma}{Lemma}{Lemmas}
\crefname{corollary}{corollary}{corollaries}
\Crefname{corollary}{Corollary}{Corollaries}

\DeclareMathOperator*{\relu}{ReLU}

\newcommand{\taps}{\textsc{TAPS}\xspace}
\newcommand{\staps}{\textsc{STAPS}\xspace}
\newcommand{\mtlibp}{\textsc{MTL-IBP}\xspace}

\newcommand{\mnbab}{\textsc{MN-BaB}\xspace}
\newcommand{\pgd}{\textsc{PGD}\xspace}

\newcommand{\crown}{\textsc{CROWN}\xspace}
\newcommand{\crownibp}{\textsc{CROWN-IBP}\xspace}

\newcommand{\deepz}{\textsc{DeepZ}\xspace}
\newcommand{\ibp}{\textsc{IBP}\xspace}
\newcommand{\sabr}{\textsc{SABR}\xspace}

\newcommand{\diffai}{\textsc{DiffAI}\xspace}

\newcommand{\ctbench}{\textsc{CTBench}\xspace}
\newcommand{\edac}{\textsc{EDAC}\xspace}
\newcommand{\rob}{\text{rob}}
\newcommand{\nat}{\text{nat}}
\newcommand{\autoattack}{\textsc{AutoAttack}\xspace}

\newcommand{\cifar}{CIFAR-10\xspace}
\newcommand{\TIN}{\textsc{TinyImageNet}\xspace}

\newcommand{\mnist}{\textsc{MNIST}\xspace}

\newcommand{\cnnf}{\texttt{CNN5}\xspace}
\newcommand{\cnns}{\texttt{CNN7}\xspace}

\usepackage{siunitx}
\newcolumntype{d}[1]{S[table-format=#1]}

\colorlet{cbackground}{c7!20}
\colorlet{cinput}{my-full-blue}
\colorlet{clatent}{my-full-orange}
\colorlet{coutput}{my-full-red}
\colorlet{cinputshape}{my-full-blue!60}
\colorlet{clatentshape}{my-full-orange!60}
\colorlet{coutputshape}{my-full-red!60}
\colorlet{cexact}{my-full-blue!80}
\colorlet{cexactlatent}{my-full-green!40}
\colorlet{cfwd}{my-full-gray}
\colorlet{cpgd}{my-full-purple!85}
\colorlet{cpgdsignle}{my-full-red!90!black!65}
\colorlet{cibp}{my-full-green!85}

\colorlet{cbwd}{my-full-red!90!black!65}
\colorlet{ctool}{c4!100}
\colorlet{netinside}{c7!100}

\tikzstyle{inputstyle}=[draw=cinputshape, opacity=1.0, ultra thick]
\tikzstyle{latentstyle}=[draw=clatentshape, opacity=1.0, ultra thick]
\tikzstyle{outputstyle}=[draw=coutputshape, opacity=1.0, ultra thick]
\tikzstyle{point}=[line width=0.5pt, draw=black, cross out, inner sep=0pt, minimum width=3pt, minimum height=3pt, anchor=center, very thick]
\tikzstyle{mid arrow}=[postaction={
    decorate,
    decoration={
            markings,
            mark=at position 0.7 with {
                    \arrow[#1]{stealth}
                }
        }}]
\tikzstyle{fwdarrow}=[color=cfwd, thick, dashed,-stealth]
\tikzstyle{pgdarrow}=[color=cpgd, thick, -stealth]
\tikzstyle{ibparrow}=[color=cibp, thick, dotted, mid arrow]

\tikzstyle{toolstyle}=[dotted, line width = 1.2pt,draw=ctool]
\tikzstyle{ibpstyle}=[dash pattern=on 5pt off 2pt, cibp]
\tikzstyle{exactlatentstyle}=[fill=cexactlatent, opacity=0.8, draw=none]
\tikzstyle{wcpoint}=[line width=0.5pt, draw=black, fill=black, circle, inner sep=0pt, minimum width=2.5pt, minimum height=2.5pt, anchor=center]
\tikzstyle{pgdsinglearrow}=[color=cpgdsignle, thick]
\tikzstyle{bwdarrow}=[color=cbwd,  line width = 1.2pt,, dotted]
\tikzstyle{pane}=[fill=cbackground, rectangle, rounded corners=2pt]

\newcommand{\newprotectedcommand}[2]{\newcommand{#1}{\protecting{#2}}}

\newprotectedcommand{\markerfwd}{\tikz[]{\draw[-stealth, fwdarrow] (0, 0) -- (0.3, 0); \node[anchor=center, minimum width=0pt, minimum height=4pt, inner sep=0pt] () at (0.15, 0.0) {};}\xspace}
\newprotectedcommand{\markerbwd}{\tikz[]{\draw[-stealth, bwdarrow] (0, 0) -- (0.3, 0); \node[anchor=center, minimum width=0pt, minimum height=4pt, inner sep=0pt] () at (0.15, 0.0) {};}\xspace}
\newprotectedcommand{\markerpgd}{\tikz[]{\draw[-stealth, pgdarrow] (0, 0) -- (0.3, 0); \node[anchor=center, minimum width=0pt, minimum height=4pt, inner sep=0pt] () at (0.15, 0.0) {};}\xspace}
\newprotectedcommand{\markerexact}{\tikz[]{\node[fill, aspect=1, inner sep=0pt, minimum size=2.1mm, exactstyle]{};}\xspace}
\newprotectedcommand{\markerexactlatent}{\tikz[]{\node[fill, aspect=1, inner sep=0pt, minimum size=2.1mm, exactlatentstyle]{};}\xspace}
\newprotectedcommand{\markeribp}{\tikz[]{\node[aspect=1, draw=cibp, inner sep=0pt, minimum size=2.1mm, ibpstyle, dash pattern=on 2.5pt off 1pt]{};}\xspace}
\newprotectedcommand{\markertool}{\tikz[]{\draw[draw=ctool, toolstyle, dash pattern=on 1pt off 0.75pt] (0,0.2) -- (0.2,0.2) -- (0.2,0);}\xspace}
\newprotectedcommand{\markerwc}{\tikz[]{\node[wcpoint] at (0,0.01) {};\node[] at (0,0) {};}\xspace}
\newprotectedcommand{\markerpoint}{\tikz[]{\node[point]{};}\xspace}
\newprotectedcommand{\markersinglepgd}{\tikz[]{\draw[pgdsinglearrow, dashed, line width =1.1pt] (0, 0) -- (0.3, 0); \node[anchor=center, minimum width=0pt, minimum height=4pt, inner sep=0pt] () at (0.15, 0.0) {};}\xspace}

\newprotectedcommand{\markermultipgd}{\tikz[]{\draw[pgdarrow, dotted, dash pattern=on 1pt off 0.75pt] (0,0.2) -- (0.2,0.2) -- (0.2,0);}\xspace}

\newprotectedcommand{\markerwcsingle}{\tikz[]{\node[point,pgdsinglearrow] at (0,0.0) {};\node[] at (0,0) {};}\xspace}
\newprotectedcommand{\markeradvpoint}{\tikz[]{\node[point,cpgd]{};}\xspace}

\newprotectedcommand{\markersabr}{\tikz[]{\node[fill, diamond, color=csabr,inner sep=0,  minimum size=2.5mm, aspect=0.5]{};}\xspace}
\newprotectedcommand{\markertaps}{\tikz[]{\node[aspect=1, fill=c4, inner sep=0pt, minimum size=2.1mm]{};}\xspace}

\definecolor{plt-purple}{HTML}{c0affb}
\definecolor{plt-green}{HTML}{5eccab}
\newprotectedcommand{\purplebar}{\tikz[]{\draw[fill=plt-purple!50!white] (0,0) -- (0,0.25) -- (0.25,0.25) -- (0.25,0) -- cycle;}\xspace}

\newprotectedcommand{\purplebarhatched}{\tikz[]{\draw[fill=plt-purple!50!white, postaction={pattern=north east lines}] (0,0) -- (0,0.25) -- (0.25,0.25) -- (0.25,0) -- cycle;}\xspace}
\newprotectedcommand{\greenbar}{\tikz[]{\draw[fill=plt-green!50!white] (0,0) -- (0,0.25) -- (0.25,0.25) -- (0.25,0) -- cycle;}\xspace}

\newprotectedcommand{\greenbarhatched}{\tikz[]{\draw[fill=plt-green!50!white, postaction={pattern=north east lines}] (0,0) -- (0,0.25) -- (0.25,0.25) -- (0.25,0) -- cycle;}\xspace}

\icmltitlerunning{\ctbench: A Library and Benchmark for Certified Training}

\begin{document}

\twocolumn[
	\icmltitle{\ctbench: A Library and Benchmark for Certified Training}

\icmlsetsymbol{equal}{*}

\begin{icmlauthorlist}
\icmlauthor{Yuhao Mao}{ethz}
\icmlauthor{Stefan Balauca}{insait}
\icmlauthor{Martin Vechev}{ethz}
\end{icmlauthorlist}

\icmlaffiliation{ethz}{Department of Computer Science, ETH Z\"urich, Switzerland}
\icmlaffiliation{insait}{INSAIT, Sofia University "St. Kliment Ohridski", Sofia, Bulgaria}

\icmlcorrespondingauthor{Yuhao Mao}{\email{yuhao.mao@inf.ethz.ch}}

\icmlkeywords{Machine Learning, ICML}

\vskip 0.3in
]

\printAffiliationsAndNotice{}  %

\begin{abstract}
 Training certifiably robust neural networks is an important but challenging task. While many algorithms for (deterministic) certified training have been proposed, they are often evaluated on different training schedules, certification methods, and systematically under-tuned hyperparameters, making it difficult to compare their performance. To address this challenge, we introduce \ctbench, a unified library and a high-quality benchmark for certified training that evaluates all algorithms under fair settings and systematically tuned hyperparameters. We show that (1) almost all algorithms in \ctbench surpass the corresponding reported performance in literature in the magnitude of algorithmic improvements, thus establishing new state-of-the-art, and (2) the claimed advantage of recent algorithms drops significantly when we enhance the outdated baselines with a fair training schedule, a fair certification method and well-tuned hyperparameters. Based on \ctbench, we provide new insights into the current state of certified training, including (1) certified models have less fragmented loss surface, (2) certified models share many mistakes, (3) certified models have more sparse activations, (4) reducing regularization cleverly is crucial for certified training especially for large radii and (5) certified training has the potential to improve out-of-distribution generalization. We are confident that \ctbench will serve as a benchmark and testbed for future research in certified training.

\end{abstract}
\section{Introduction}

As a crucial component of trustworthy artificial intelligence, adversarial robustness \citep{SzegedyZSBEGF13,GoodfellowSS14}, \ie, resilience to small input perturbations, has established itself as an important research area. While initially the community focused on heuristic methods to craft adversarial examples and defenses against them, it turned out that such defenses are often brittle and can be evaded by adaptive adversaries \citep{AthalyeC018,TramerCBM20}. Thus, neural network certification has emerged as a method for providing provable guarantees on the robustness of a given network \citep{GehrMDTCV18,WongK18,ZhangWCHD18,SinghGPV19}.

Two families of neural network certification methods have been proposed: complete methods  \citep{KatzBDJK17,TjengXT19} which compute the exact bounds but are extremely computationally expensive, and convex-relaxation based methods \citep{ZhangWCHD18,SinghGPV19} which are more scalable but provide approximate bounds. State-of-the-art (SOTA) verifiers \citep{XuZ0WJLH21,FerrariMJV22,ZhangWXLLJ22} combine both approaches, by using convex relaxations to speed up the solving of complete methods via Branch-and-Bound \citep{BunelLTTKK20}.

However, the scalability of neural network certification is still a major challenge since the computational complexity of SOTA verifiers grows exponentially with network size. To tackle this issue, certified training \citep{MirmanGV18,GowalIBP2018} was proposed to train neural networks that are amenable to certification. Such methods are typically categorized into two groups: (1) training with a sound upper bound of the robust loss \citep{GowalIBP2018,ZhangCXGSLBH20,ShiWZYH21}, and (2) training with an unsound surrogate loss that approximates the exact robust loss \citep{MuellerEFV22,MaoM0V23,palma2024expressive}. The latter group has been shown to be more effective.

While certified training has made significant advances, there is currently no benchmark that can be used to fairly evaluate the effectiveness of the different certified training methods. Specifically, the literature often compares against previous methods using quoted numbers due to high computational costs, although the verifier and certification budget differ. These unfair comparisons ultimately hinder the community from drawing reasonable conclusions on the effectiveness of certified training methods. In addition, existing works systematically under-tune hyperparameters, in order to show effectiveness against baselines, thus establishing a weaker SOTA. Further, there is no unified codebase for these methods, making future development and comparison difficult.

\begin{figure}
    \centering
    \includegraphics[width=.95\linewidth]{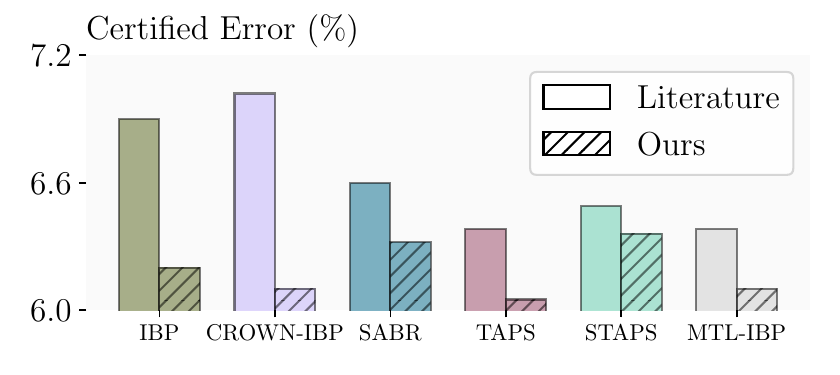}
    \caption{Reduction in certified error on \mnist $\epsilon=0.3$ (lower is better).}
    \label{tb:intro}
\end{figure} 

\paragraph{This work: a Unified Library and High-quality Benchmark for Certified Training} We address these challenges, for the first time unifying SOTA certified training methods into a single codebase called \ctbench. This enables a fair comparison between certified training methods and re-establishes a much stronger SOTA by fixing problematic implementations and systematically tuning hyperparameters. As shown in \cref{tb:intro}, these steps lead to significant improvements uniformly.
In addition, we show that the claimed advantage of recent SOTA reduces significantly when we apply the same budget and hyperparameter tuning to all methods. Based on our released model checkpoints, we provide an extensive analysis of the model properties, highlighting many new insights on its loss landscape, mistake patterns, regularization strength, model utilization, and out-of-distribution generalization. We are confident that \ctbench will serve as a benchmark and testbed for future work in certified training.

\vspace{-1mm}
\section{Related Work}
We now briefly review works mostly related to ours.

\paragraph{Benchmarking Certified Robustness} \citet{LiXL23} provides the first benchmark for certified robustness, covering not only deterministic certified training but also randomized certified training and certification methods. However, it is outdated and thus provides little insight into the current SOTA methods. For example, it reports 89\% and 51\% best certified accuracy for \mnist $\epsilon=0.3$ and \cifar $\epsilon=\frac{2}{255}$ in its benchmark, respectively, while recent methods have achieved more than 93\% and 62\% \citep{MuellerEFV22, MaoM0V23,palma2024expressive}.

\paragraph{Certified Training} \diffai \citep{MirmanGV18} and \ibp \citep{GowalIBP2018} apply box relaxation to upper bound the worst-case loss for training. Efforts have been made towards applying more precise approximations: \citet{WongSMK18} and \citet{BalunovicV20} apply \deepz \citep{SinghGMPV18}, and \citet{ZhangCXGSLBH20} incorporate linear relaxations \citep{ZhangWCHD18,SinghGPV19}. While these approximations are more precise \citep{baader2024expressivity,mao2025expressiveness}, they often lead to worse training results, attributed to non-smoothness \citep{LeeLPL21}, discontinuity and sensitivity \citep{jovanovic2022paradox} of the loss surface. Some recent work \citep{balauca2024overcoming} aims to mitigate these problems, however, the most effective training approximation is still the least precise box relaxation. In this regard, the focus of the community has shifted towards improving \ibp: \citet{ShiWZYH21} propose a new regularization and initialization paradigm to speed up \ibp training; \citet{PalmaIBPR22} apply \ibp regularization to make adversarial training certifiable; \citet{MuellerEFV22}, \citet{MaoM0V23} and \citet{palma2024expressive} propose unsound but more effective \ibp-based surrogate losses for training; \citet{mao2024understanding} propose to use wider models instead of deeper models for \ibp-based methods. These methods achieve universal advantages over non-\ibp-based methods, and are thus the focus of our work.

\vspace{-1mm}
\section{Background}

We now introduce the necessary background for our work, both concepts and training algorithms.

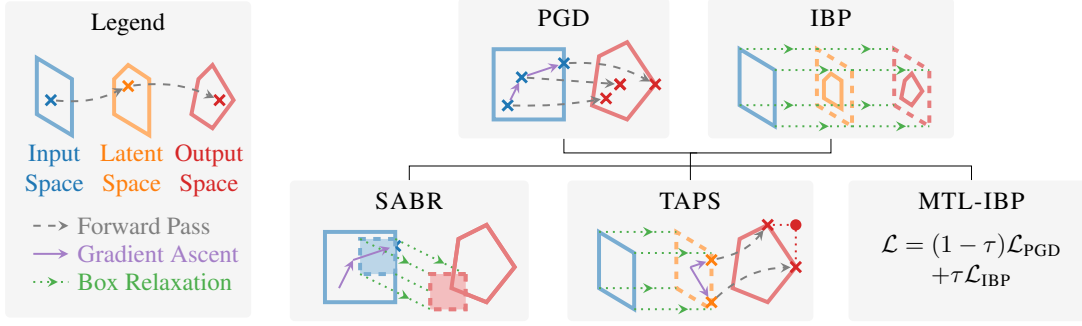
\begin{figure*}
    \centering
    \def\sep{0.05cm}
    \resizebox{.85\linewidth}{!}{
    \begin{tikzpicture}
        \begin{scope}
            \node[pane, minimum width=3.5cm, minimum height=4.5cm] (legend pane) at (0, 0) {};
            \node[align=center,anchor=north] at ($(legend pane.north)-(0,\sep)$) (legend) {Legend};
            \coordinate (input space) at ($(legend.south)+(-1.3cm,-1cm)$);
            \draw[inputstyle] (input space) -- ++(0.5cm, -0.3cm) -- ++ (0cm, 0.8cm) -- ++ (-0.5cm, 0.3cm) -- cycle;
            \node[cinput,align=center] at ($(input space)+(0.25cm,-0.8cm)$) {Input\\Space};
            \coordinate (latent space) at ($(legend.south)+(-0.2cm,-1cm)$);
            \draw[latentstyle] (latent space) -- ++(0.5cm, -0.3cm) -- ++ (0cm, 0.8cm) -- ++ (-0.3cm, 0.2cm) -- ++(-0.2cm,-0.2cm) -- cycle;
            \node[clatent, align=center] at ($(latent space)+(0.25cm,-0.8cm)$) (latent text) {Latent\\Space};
            \coordinate (output space) at ($(legend.south)+(0.9cm,-1cm)$);
            \draw[outputstyle] (output space)  -- ++(0.3cm, -0.2cm) -- ++ (0.3cm, 0.4cm) -- ++ (-0.3cm, 0.5cm) -- ++(-0.2cm,-0.2cm) -- cycle;
            \node[coutput,align=center] at ($(output space)+(0.25cm,-0.8cm)$) {Output\\Space};
            \node[point,cinput] (input point) at ($(input space)+(0.2cm, 0.2cm)$) {};
            \node[point,clatent] (latent point) at ($(latent space)+(0.2cm, 0.4cm)$) {};
            \node[point,coutput] (output point) at ($(output space)+(0.4cm, 0.2cm)$) {};
            \draw[fwdarrow] (input point) edge[bend right=15] (latent point);
            \draw[fwdarrow] (latent point) edge[bend left=15] (output point);
            \draw[fwdarrow] ($(latent text.south)+(-1.4cm,-0.3cm)$) -- ++(0.5cm, 0cm) node[right] {Forward Pass};
            \draw[pgdarrow] ($(latent text.south)+(-1.4cm,-0.7cm)$) -- ++(0.5cm, 0cm) node[right] {Gradient Ascent};
            \draw[ibparrow] ($(latent text.south)+(-1.4cm,-1.1cm)$) -- ++(0.5cm, 0cm) node[right] {Box Relaxation};
        \end{scope}

        \begin{scope}[xshift=6.2cm, yshift=1.3cm]
            \node[pane, minimum width=3cm, minimum height=2cm] (PGD pane) at (0, 0) {};
            \node[align=center,anchor=north] at ($(PGD pane.north)-(0,\sep)$) (PGD) {\pgd};
            \coordinate (input space) at ($(PGD.south)+(-1cm,-1.2cm)$);
            \draw[inputstyle] (input space) rectangle ++(1cm, 1cm);
            \coordinate (output space) at ($(PGD.south)+(0.4cm,-1cm)$);
            \draw[outputstyle] (output space)  -- ++(0.5cm, -0.2cm) -- ++ (0.4cm, 0.5cm) -- ++ (-0.4cm, 0.6cm) -- ++(-0.4cm,-0.2cm) -- cycle;
            \node[point,cinput] (input point) at ($(input space)+(0.2cm, 0.2cm)$) {};
            \node[point,coutput] (output point) at ($(output space)+(0.2cm, 0.1cm)$) {};
            \draw[fwdarrow] (input point) edge[bend right=5] (output point);
            \node[point,cinput] (input point 1) at ($(input point)+(0.2cm, 0.4cm)$) {};
            \draw[pgdarrow] (input point) -- (input point 1);
            \node[point,coutput] (output point 1) at ($(output point)+(0.2cm, 0.2cm)$) {};
            \draw[fwdarrow] (input point 1) -- (output point 1);
            \node[point,cinput] (input point 2) at ($(input space)+(1cm, 0.8cm)$) {};
            \draw[pgdarrow] (input point 1) -- (input point 2);
            \node[point,coutput] (output point 2) at ($(output space)+(0.9cm, 0.3cm)$) {};
            \draw[fwdarrow] (input point 2) edge[bend left=15] (output point 2);
        \end{scope}

        \begin{scope}[xshift=10cm, yshift=1.3cm]
            \node[pane, minimum width=3.5cm, minimum height=2cm] (IBP pane) at (0, 0) {};
            \node[align=center,anchor=north] at ($(IBP pane.north)-(0,\sep)$) (IBP) {\ibp};
            \coordinate (input space) at ($(IBP.south)+(-1.3cm,-1cm)$);
            \draw[inputstyle] (input space) -- ++(0.5cm, -0.3cm) -- ++ (0cm, 0.8cm) -- ++ (-0.5cm, 0.3cm) -- cycle;
            \coordinate (latent space) at ($(IBP.south)+(-0.2cm,-1cm)$);
            \draw[latentstyle] ($(latent space)+(0.1cm,0.1cm)$) -- ++(0.25cm, -0.15cm) -- ++ (0cm, 0.4cm) -- ++ (-0.15cm, 0.1cm) -- ++(-0.1cm,-0.1cm) -- cycle;
            \draw[latentstyle, dashed] (latent space) -- ++(0.5cm, -0.3cm) -- ++ (0cm, 0.8cm) -- ++ (-0.5cm, 0.3cm) -- cycle;
            \draw[ibparrow] (input space) -- (latent space);
            \draw[ibparrow] ($(input space)+(0.5cm,-0.3cm)$) -- ($(latent space)+(0.5cm,-0.3cm)$);
            \draw[ibparrow] ($(input space)+(0.5cm,0.5cm)$) -- ($(latent space)+(0.5cm,0.5cm)$);
            \draw[ibparrow] ($(input space)+(0cm,0.8cm)$) -- ($(latent space)+(0cm,0.8cm)$);
            \coordinate (output space) at ($(IBP.south)+(0.9cm,-1cm)$);
            \draw[outputstyle] ($(output space)+(0.1cm,0.1cm)$)  -- ++(0.15cm, -0.1cm) -- ++ (0.15cm, 0.2cm) -- ++ (-0.15cm, 0.25cm) -- ++(-0.1cm,-0.1cm) -- cycle;
            \draw[outputstyle, dashed] (output space) -- ++(0.5cm, -0.3cm) -- ++ (0cm, 0.8cm) -- ++ (-0.5cm, 0.3cm) -- cycle;
            \draw[ibparrow] (latent space) -- (output space);
            \draw[ibparrow] ($(latent space)+(0.5cm,-0.3cm)$) -- ($(output space)+(0.5cm,-0.3cm)$);
            \draw[ibparrow] ($(latent space)+(0.5cm,0.5cm)$) -- ($(output space)+(0.5cm,0.5cm)$);
            \draw[ibparrow] ($(latent space)+(0cm,0.8cm)$) -- ($(output space)+(0cm,0.8cm)$);
        \end{scope}

        \begin{scope}[xshift=4cm, yshift=-1.3cm]
            \node[pane, minimum width=3.4cm, minimum height=2cm] (SABR pane) at (0, 0) {};
            \node[align=center,anchor=north] at ($(SABR pane.north)-(0,\sep)$) (SABR) {\sabr};
            \coordinate (input space) at ($(SABR.south)+(-1.2cm,-1.2cm)$);
            \draw[inputstyle] (input space) rectangle ++(1cm, 1cm);
            \coordinate (output space) at ($(SABR.south)+(0.6cm,-1cm)$);
            \draw[outputstyle] (output space)  -- ++(0.5cm, -0.2cm) -- ++ (0.4cm, 0.5cm) -- ++ (-0.4cm, 0.6cm) -- ++(-0.4cm,-0.2cm) -- cycle;
            \coordinate (input point) at ($(input space)+(0.2cm, 0.2cm)$) {};
            \coordinate (input point 1) at ($(input point)+(0.2cm, 0.4cm)$) {};
            \node[point,coutput] (output point 1) at ($(output point)+(0.2cm, 0.2cm)$) {};
            \node[point,cinput] (input point 2) at ($(input space)+(1cm, 0.8cm)$) {};
            \coordinate (small input box) at ($(input point 2)+(0cm,0.1cm)$);
            \draw[inputstyle, dashed, fill=cinput!30] (small input box) rectangle ++(-0.5cm, -0.5cm);
            \draw[pgdarrow] (input point) -- (input point 1);
            \draw[pgdarrow] (input point 1) -- (input point 2);
            \coordinate (output point 2) at ($(output space)+(0.9cm, 0.3cm)$) {};
            \coordinate (small output box) at ($(output space)+(0.2cm,0.2cm)$);
            \draw[outputstyle, dashed, fill=coutput!30] (small output box) rectangle ++(-0.5cm, -0.5cm);
            \draw[ibparrow] (small input box) -- (small output box);
            \draw[ibparrow] ($(small input box)-(0.5cm,0.5cm)$) -- ($(small output box)-(0.5cm,0.5cm)$);
            \draw[ibparrow] ($(small input box)-(0.5cm,0cm)$) -- ($(small output box)-(0.5cm,0cm)$);
            \draw[ibparrow] ($(small input box)-(0cm,0.5cm)$) -- ($(small output box)-(0cm,0.5cm)$);
            \draw[outputstyle, dashed, fill=coutput!30] (small output box) rectangle ++(-0.5cm, -0.5cm);
            \draw[outputstyle] (output space)  -- ++(0.5cm, -0.2cm) -- ++ (0.4cm, 0.5cm) -- ++ (-0.4cm, 0.6cm) -- ++(-0.4cm,-0.2cm) -- cycle;
        \end{scope}

        \begin{scope}[xshift=8cm, yshift=-1.3cm]
            \node[pane, minimum width=3.5cm, minimum height=2cm] (TAPS pane) at (0, 0) {};
            \node[align=center,anchor=north] at ($(TAPS pane.north)-(0,\sep)$) (TAPS) {\taps};
            \coordinate (input space) at ($(TAPS.south)+(-1.3cm,-1cm)$);
            \draw[inputstyle] (input space) -- ++(0.5cm, -0.3cm) -- ++ (0cm, 0.8cm) -- ++ (-0.5cm, 0.3cm) -- cycle;
            \coordinate (latent space) at ($(TAPS.south)+(-0.2cm,-1cm)$);
            \draw[latentstyle, dashed] (latent space) -- ++(0.5cm, -0.3cm) -- ++ (0cm, 0.8cm) -- ++ (-0.5cm, 0.3cm) -- cycle;
            \draw[ibparrow] (input space) -- (latent space);
            \draw[ibparrow] ($(input space)+(0.5cm,-0.3cm)$) -- ($(latent space)+(0.5cm,-0.3cm)$);
            \draw[ibparrow] ($(input space)+(0.5cm,0.5cm)$) -- ($(latent space)+(0.5cm,0.5cm)$);
            \draw[ibparrow] ($(input space)+(0cm,0.8cm)$) -- ($(latent space)+(0cm,0.8cm)$);
            \coordinate (output space) at ($(TAPS.south)+(0.6cm,-1cm)$);
            \draw[outputstyle] (output space)  -- ++(0.5cm, -0.2cm) -- ++ (0.4cm, 0.5cm) -- ++ (-0.4cm, 0.6cm) -- ++(-0.4cm,-0.2cm) -- cycle;
            \coordinate (latent point) at ($(latent space)+(0.2cm, 0.3cm)$);
            \node[point, clatent] (latent point 1) at ($(latent point)+(0.3cm, 0.1cm)$) {};
            \draw[pgdarrow] (latent point) -- (latent point 1);
            \node[point, coutput] (output point 1) at ($(output space)+(0.5cm, 0.9cm)$) {};
            \draw[fwdarrow] (latent point 1) edge[bend right=20] (output point 1);
            \node[point, clatent] (latent point 2) at ($(latent point)+(0.3cm, -0.5cm)$) {};
            \draw[pgdarrow] (latent point) -- (latent point 2);
            \node[point, coutput] (output point 2) at ($(output space)+(0.9cm, 0.3cm)$) {};
            \draw[fwdarrow] (latent point 2) edge[bend left=20] (output point 2);
            \draw[coutput, dotted, thick] (output point 1) -| (output point 2);
            \fill[coutput] (output point 1 -| output point 2) circle (0.08cm);
        \end{scope}

        \begin{scope}[xshift=12cm, yshift=-1.3cm]
            \node[pane, minimum width=3.5cm, minimum height=2cm] (MTL-IBP pane) at (0, 0) {};
            \node[align=center,anchor=north] at ($(MTL-IBP pane.north)-(0,\sep)$) (MTL-IBP) {\mtlibp};
            \node[align=center] at ($(MTL-IBP.south)+(-0cm,-0.6cm)$) {$\gL=(1-\tau)\gL_{\pgd}$\\$+\tau \gL_{\ibp}$};
        \end{scope}

        \coordinate (join point upper) at ($(PGD pane.south)!.5!(IBP pane.south)+(0cm, -0.2cm)$);
        \coordinate (join point lower) at ($(TAPS pane.north)+(0cm, 0.2cm)$);
        \draw (join point upper) -| (join point lower);
        \draw (PGD pane.south) |- (join point upper);
        \draw (IBP pane.south) |- (join point upper);
        \draw (SABR pane.north) |- (join point lower);
        \draw (TAPS pane.north) |- (join point lower);
        \draw (MTL-IBP pane.north) |- (join point lower);

    \end{tikzpicture}
    }
    \caption{Conceptual overview of core algorithms built into \ctbench.} \label{fig:overview}
\end{figure*}

\subsection{Training for Robustness}

We present the mathematical notations on adversarial and certified training here. We consider a neural network classifier $f_\theta(x)$ that estimates the log-probability of each class and predicts the class with the highest estimated log-probability.

\paragraph{Adversarial Training} A classifier $f_\theta(x)$ is said to be \emph{adversarially robust} with radius $\epsilon$ w.r.t. $L_p$ perturbation if $f_\theta(x+\delta)=y$ for all $\|\delta\|_p \le \epsilon$, where $y$ is the ground truth label of $x$.  Finding an adversarially robust classifier is formally defined to solve a min-max problem $\theta = \argmin_\theta \E_{x,y} \max_{\|\delta\|_p\le \epsilon} L(x+\delta)$. In this regard, adversarial training solves the inner maximization problem by generating adversarial examples during training, and the outer minimization problem by optimizing the empirical loss of adversarial examples.

\paragraph{Certified Training} A classifier $f_\theta(x)$ is said to be \emph{certifiably robust} if it is adversarially robust and there exists a sound verifier that certifies the robustness. A verifier typically computes an upper bound on the margin $f_i(x+\delta)-f_y(x+\delta)$ and certifies its robustness if the upper bound is negative for all $i\ne y$. Certified training thus replaces the inner maximization problem with an upper bound and minimizes the upper bound during training instead. Since existing certified training algorithms focus solely on $L_\infty$ distance, we only consider $L_\infty$ perturbations in this work and omit the distance type in the notation.

\paragraph{Metrics} The main metric for certified training is \emph{certified accuracy}, defined to be the ratio of certifiably robust samples in the dataset; \emph{certified error} is defined as one minus the certified accuracy. The ratio of correctly classified samples in the dataset is thus called \emph{natural accuracy}. For reference, we include \emph{adversarial accuracy} as well, defined to be the ratio of adversarially robust samples in the dataset. We apply one of the most widely used SOTA certification methods, \mnbab \citep{FerrariMJV22}, as the verifier. To compute adversarial accuracy, we apply the strong \autoattack \citep{Croce020a} for adversarially trained models, and a combination of \pgd attack and branch-and-bound attack from \mnbab for certifiably trained models. Both attacks have similar empirical strengths, with the latter being slightly stronger on models trained by certified training algorithms due to the completeness of the branch-and-bound attack.

\subsection{Algorithms in \ctbench} \label{sec:algorithms}

Now we briefly introduce the core algorithms built into \ctbench. Concepts behind them are visualized in \cref{fig:overview}. A theoretical complexity analysis is provided in \cref{tab:complexity} in \cref{app:computation}.

\paragraph{\pgd and \edac} Projected Gradient Descent (\pgd)~\citep{MadryMSTV18} is the most widely recognized adversarial training method. Starting from a randomly initialized point, \pgd solves the inner maximization problem by iteratively taking a step towards the gradient ascent direction and clipping the solution into the valid perturbation set. Then, it uses the generated adversarial input $x^\prime$ to lower bound the worst case loss as $L(x^\prime)$. \citet{Croce020a} find that \pgd-trained models remains effective against strong attacks, thus it is popular as an integrated part of many certified training methods \citep{MuellerEFV22,MaoM0V23,palma2024expressive}. To further improve adversarial robustness, \citet{zhang2023generating} improves adversarial generalization via an extra-gradient method called \edac, which remains one of the SOTA methods in adversarial training. These methods achieve good but uncertifiable adversarial robustness, hence we use them as adversarial robustness baselines.

\paragraph{\ibp} Interval Bound Propagation (\ibp) \citep{MirmanGV18,GowalIBP2018} uses interval analysis to approximate the output range of each layer. For example, for the toy network $y = 2-\relu(x_1 + x_2)$ with input bounds $x_1,x_2\in[-1,1]$, it first computes the output range of the first layer as $x_1 + x_2 \in [-1,1]+[-1,1] \subseteq [-2,2]$, the second layer as $\relu([-2,2]) \subseteq [0,2]$ and then final layer as $2-[0,2]\subseteq [0,2]$, thus proving $y\ge 0$ for all possible $x_1,x_2\in[-1,1]$. Similarly, IBP computes the layer-wise bounds and then derives an upper bound of the worst-case loss based on the output bounds of the final layer. To stably train models with \ibp, \citet{ShiWZYH21} propose to rescale the parameter initialization to ensure constant growth of \ibp bounds and a specialized regularization to control the activation status of neurons. They also show that adding a batch norm \citep{IoffeS15} layer before every ReLU layer can improve \ibp training. These training tricks are adopted by every \ibp-based method introduced below. For brevity, we refer to this variant as \ibp in the rest of the paper unless otherwise stated, since it improves the original \ibp universally with tricks that facilitate training.

\paragraph{\crownibp} \crownibp \citep{ZhangCXGSLBH20} tightens the imprecise interval analysis with linear relaxations of ReLU layers based on \ibp bounds and only solves the linear constraints for the final layer output based on \crown \citep{ZhangWCHD18}, avoiding prohibitive costs during training. To further reduce the cost of solving the bounds for each class, \citet{XuS0WCHKLH20} propose a loss fusion trick to only solve for the final loss, thus reducing the asymptotic complexity by a factor equal to the number of classes. For brevity, we refer to this variant as \crownibp in the rest of the paper unless otherwise stated, since the original \crownibp cannot scale to datasets with many classes, such as \TIN.

\paragraph{\sabr} Since \ibp is often criticized for the increasingly strong regularization w.r.t. input radius imposed on the neural network, \sabr \citep{MuellerEFV22} proposes to use \ibp only for a carefully chosen small box inside the original input box for \ibp training. More specifically, it first conducts a \pgd attack in the input domain to find an approximately worst-case input, and then takes the surrounding small box with radius $\lambda \epsilon$ around the found input as the input box for \ibp training, where $\lambda$ is a pre-defined ratio. For exceptional cases (specifically \cifar $\epsilon=\frac{2}{255}$), \sabr further shrinks the output box of every ReLU towards zero by a pre-defined constant to further reduce the regularization.

\paragraph{\taps and \staps} Observing that \ibp relaxation error grows exponentially w.r.t. model depth \citep{MuellerEFV22,mao2024understanding}, \taps \citep{MaoM0V23} proposes to split the network into two subparts, using \ibp for the first subpart and \pgd for the other. This way, the over-approximation from \ibp and the under-approximation from \pgd partially cancel out, yielding a more precise approximation of the worst-case loss. Further, \taps uses a separate \pgd attack to estimate the bounds of every class to align better with the certification objective. \staps \citep{MaoM0V23} combines \taps with \sabr by using the adversarial small box for \taps training, thus further reducing regularization.

\paragraph{\mtlibp} \citet{palma2024expressive} formalizes a family of surrogate loss functions that interpolate between \pgd and \ibp training. We study \mtlibp, one of the most effective algorithms in this family. \mtlibp linearly interpolates between \pgd loss and \ibp loss, \ie, $\gL=(1-\tau)\gL_{\pgd}+\tau \gL_{\ibp}$, where $\tau$ is the pre-defined \ibp coefficient. To allow more fine-grained control of the interpolation, \mtlibp uses a larger input radius for the \pgd attack for \cifar when $\epsilon=\frac{2}{255}$.

\section{A Unified Library and High-quality Benchmark for Certified Training}
We now discuss \ctbench, both the unified library and the corresponding benchmark.

\subsection{The \ctbench library} \label{sec:implementation}

\begin{table*}[t]
    \centering
    \caption{\ctbench results with comparison to the literature. We include the natural accuracy of standard training with \cnns on each dataset for reference. The best numbers are in bold and those exceeding the literature results are underlined.}
    \label{tab:ctbench}
    \renewcommand{\arraystretch}{1.0}
    \begin{adjustbox}{width=.87\linewidth,center}
        \begin{threeparttable}
            \begin{tabular}{ccccccccc}
                \toprule
                Dataset                  & \multirow{2}*{$\epsilon_\infty$} & \multirow{2}*{Training Method} & \multirow{2}*{Source}       & \multicolumn{2}{c}{Nat. [\%]} & \multicolumn{2}{c}{Cert. [\%]} & Adv. [\%]                                              \\
                Std. Nat. [\%]          &                                  &                                &                             & Literature                    & \ctbench                       & Literature     & \ctbench                   & \ctbench \\
                \midrule
                \multirow[b]{8}*{\mnist} & \multirow{8}*{0.1}               & \pgd                           & /                           & /                             & 99.47                          & /              & $\approx\!0^\dagger$       & 98.97    \\
                                         &                                  & \edac                          & /                           & /                             & 99.58                          & /              & $\approx\!0^\dagger$       & 98.95    \\
                \cmidrule[.1pt](rl){3-3}
                                         &                                  & \ibp                           & \citet{ShiWZYH21}           & 98.84                         & \underline{98.87}              & 97.95          & \underline{98.26}          & 98.27    \\
                                         &                                  & \crownibp                      & \citet{XuS0WCHKLH20}        & 98.83                         & \underline{98.94}              & 97.76          & \underline{98.21}          & 98.23    \\
                                         &                                  & \sabr                          & \citet{MuellerEFV22}        & 99.23                         & 99.08                          & 98.22          & \underline{98.40}          & 98.47    \\
                                         &                                  & \taps                          & \citet{MaoM0V23}            & 99.19                         & 99.16                          & \textbf{98.39} & \textbf{\underline{98.52}} & 98.58    \\
                                         &                                  & \staps                         & \citet{MaoM0V23}            & 99.15                         & 99.11                          & 98.37          & \underline{98.47}          & 98.50    \\
                                         &                                  & \mtlibp                        & \citet{palma2024expressive} & \textbf{99.25 }               & \textbf{99.18}                 & 98.38          & 98.37                      & 98.44    \\
                \cmidrule(rl){2-9}
                \multirow[t]{8}*{99.50}  & \multirow{8}*{0.3}               & \pgd                           & /                           & /                             & 99.43                          & /              & $\approx\!0^\dagger$       & 93.83    \\
                                         &                                  & \edac                          & /                           & /                             & 99.51                          & /              & $\approx\!0^\dagger$       & 95.02    \\
                \cmidrule[.1pt](rl){3-3}
                                         &                                  & \ibp                           & \citet{ShiWZYH21}           & 97.67                         & \underline{98.54}              & 93.10          & \underline{93.80}          & 94.30    \\
                                         &                                  & \crownibp                      & \citet{XuS0WCHKLH20}        & 98.18                         & \underline{98.48}              & 92.98          & \underline{93.90}          & 94.29    \\
                                         &                                  & \sabr                          & \citet{MuellerEFV22}        & 98.75                         & 98.66                          & 93.40          & \underline{93.68}          & 94.46    \\
                                         &                                  & \taps                          & \citet{MaoM0V23}            & 97.94                         & \underline{98.56}              & \textbf{93.62} & \textbf{\underline{93.95}} & 94.66    \\
                                         &                                  & \staps                         & \citet{MaoM0V23}            & 98.53                         & \textbf{\underline{98.74}}     & 93.51          & \underline{93.64}          & 94.36    \\
                                         &                                  & \mtlibp                        & \citet{palma2024expressive} & \textbf{98.80}                & \textbf{98.74}                 & \textbf{93.62} & \underline{93.90}          & 94.55    \\
                \cmidrule(rl){1-9}
                \multirow[b]{8}*{\cifar} & \multirow{8}*{$\frac{2}{255}$}   & \pgd                           & /                           & /                             & 88.67                          & /              & $\approx\!0^\dagger$       & 72.41    \\
                                         &                                  & \edac                          & /                           & /                             & 89.18                          & /              & $\approx\!0^\dagger$       & 72.42    \\
                \cmidrule[.1pt](rl){3-3}
                                         &                                  & \ibp                           & \citet{ShiWZYH21}           & 66.84                         & \underline{67.49}              & 52.85          & \underline{55.99}          & 56.10    \\
                                         &                                  & \crownibp                      & \citet{XuS0WCHKLH20}        & 71.52                         & 67.60                          & 53.97          & \underline{57.11}          & 57.28    \\
                                         &                                  & \sabr                          & \citet{MuellerEFV22}        & 79.24                         & 77.86                          & 62.84          & \underline{63.61}          & 65.56    \\
                                         &                                  & \taps                          & \citet{MaoM0V23}            & 75.09                         & 74.44                          & 61.56          & 61.27                      & 62.62    \\
                                         &                                  & \staps                         & \citet{MaoM0V23}            & 79.76                         & 77.05                          & 62.98          & \underline{64.21}          & 66.09    \\
                                         &                                  & \mtlibp                        & \citet{palma2024expressive} & \textbf{80.11}                & \textbf{78.82}                 & \textbf{63.24} & \textbf{\underline{64.41}} & 67.69    \\
                \cmidrule(rl){2-9}
                \multirow[t]{8}*{91.27}  & \multirow{8}*{$\frac{8}{255}$}   & \pgd                           & /                           & /                             & 78.71                          & /              & $\approx\!0^\dagger$       & 35.93    \\
                                         &                                  & \edac                          & /                           & /                             & 78.95                          & /              & $\approx\!0^\dagger$       & 42.48    \\
                \cmidrule[.1pt](rl){3-3}
                                         &                                  & \ibp                           & \citet{ShiWZYH21}           & 48.94                         & 48.51                          & 34.97          & \underline{35.28}          & 35.48    \\
                                         &                                  & \crownibp                      & \citet{XuS0WCHKLH20}        & 46.29                         & \underline{48.25}              & 33.38          & 32.59                      & 32.77    \\
                                         &                                  & \sabr                          & \citet{MuellerEFV22}        & 52.38                         & \underline{52.71}              & 35.13          & \underline{35.34}          & 36.11    \\
                                         &                                  & \taps                          & \citet{MaoM0V23}            & 49.76                         & \underline{49.96}              & 35.10          & \underline{35.25}          & 35.69    \\
                                         &                                  & \staps                         & \citet{MaoM0V23}            & 52.82                         & 51.49                          & 34.65          & \underline{35.11}          & 35.54    \\
                                         &                                  & \mtlibp                        & \citet{palma2024expressive} & \textbf{53.35}                & \textbf{\underline{54.28}}     & \textbf{35.44} & \textbf{35.41}             & 36.02    \\
                \cmidrule(rl){1-9}
                \multirow[b]{4}*{\TIN}   & \multirow{8}*{$\frac{1}{255}$}   & \pgd                           & /                           & /                             & 46.78                          & /              & $\approx\!0^\dagger$       & 33.16    \\
                                         &                                  & \edac                          & /                           & /                             & 46.79                          & /              & $\approx\!0^\dagger$       & 33.16    \\
                \cmidrule[.1pt](rl){3-3}
                                         &                                  & \ibp                           & \citet{ShiWZYH21}           & 25.92                         & \underline{26.77}              & 17.87          & \underline{19.82}          & 19.84    \\
                                         &                                  & \crownibp                      & \citet{XuS0WCHKLH20}        & 25.62                         & \underline{28.44}              & 17.93          & \underline{22.14}          & 22.31    \\
                \multirow[t]{4}*{47.96}  &                                  & \sabr                          & \citet{MuellerEFV22}        & 28.85                         & \underline{30.58}              & 20.46          & \underline{20.96}          & 21.16    \\
                                         &                                  & \taps                          & \citet{MaoM0V23}            & 28.34                         & \underline{28.64}              & 20.82          & \underline{21.58}          & 21.71    \\
                                         &                                  & \staps                         & \citet{MaoM0V23}            & 28.98                         & \underline{30.63}              & 22.16          & \underline{22.31}          & 22.57    \\
                                         &                                  & \mtlibp                        & \citet{palma2024expressive} & \textbf{37.56}                & \textbf{35.97}                 & \textbf{26.09} & \textbf{\underline{27.73}} & 28.49    \\
                \bottomrule
            \end{tabular}
            \begin{tablenotes}
                \item $\dagger$ None of the first 10 samples are certified due to the time limit of 1000 seconds per sample.
            \end{tablenotes}
        \end{threeparttable}
    \end{adjustbox}
    \vspace{-4mm}
\end{table*}

We implement every algorithm described in \cref{sec:algorithms} in a unified framework. The training loss is composed of three components: the natural loss which measures performance on clean inputs, the robust loss which measures robust performance depending on the concrete algorithms and regularization losses which are used to stabilize training and improve generalization. Formally, the training loss is defined as $\gL = (1-w_{\rob})\gL_{\nat} + w_{\rob} \gL_{\rob} + \gL_{\text{reg}}$. We mainly use $L_1$ regularization to reduce overfitting and the warmup regularization proposed by \citet{ShiWZYH21} to improve certified training methods. The \ibp initialization \citep{ShiWZYH21} is applied for every certified training method, while adversarial training is initialized with Kaiming uniform \citep{HeZRS15}. Every method has a warmup phase where $\epsilon$ is increased from $0$ to the target value and a fine-tuning phase where the model continues to train at the targeted $\epsilon$ to converge. The learning rate is held constant during the warmup phase and decayed twice in the fine-tuning phase with a constant multiplier. We use \cnns as the model architecture, in agreement with recent literature \citep{ShiWZYH21,MuellerEFV22,MaoM0V23,palma2024expressive}.

Due to the importance of batch norm in certified training, we consider it as a native part of \ctbench. Specifically, the best practice so far is to set batch norm statistics based on the clean input and use this for computing \ibp bounds. However, we find several problematic implementations of batch norm in the literature: (1) when gradient accumulation is involved, the batch norm statistics are not updated correctly, as sub-batch statistics are applied for training; (2) batch norm statistics change more than once before taking a gradient step, as typically the exponentially accumulated statistics are used for conducting a \pgd attack and thus evaluating $\gL_{\rob}$, while $\gL_{\nat}$ is evaluated with batch statistics. The first problem makes gradient accumulation ineffective since the quality of batch statistics depends highly on the batch size, and the second problem prevents training with $w_{\rob} \in (0,1)$ due to the varied parameters. To address the first problem, we propose to use full batch statistics during gradient accumulation, which leads to slim overheads but allows arbitrary gradient accumulation, as a forward pass is usually much cheaper than a full batch update in certified training. To address the second problem, we conduct \pgd attacks with the batch statistics as well and evaluate everything with the current batch statistics. This way, the batch norm statistics are set once per batch just like standard training, allowing training with the combination of $\gL_{\nat}$ and $\gL_{\rob}$.  We remark that the identified problems are systematically ignored in the literature, thus may only be discovered by carefully reading the implementations, which is infeasible for most researchers.

In addition, we find that models trained with the hyperparameters reported in the literature frequently show strong overfitting patterns. To remediate this, we conduct a magnitude search for $L_1$ regularization until the train and validation performance roughly match. To further aid generalization, we apply Stochastic Weight Averaging  \citep{IzmailovPGVW18} for methods that cannot provide metrics for model selection, e.g., \mtlibp. A more detailed description of the implementation can be found in \cref{app:exp_setting}.

\subsection{The \ctbench benchmark}

\cref{tab:ctbench} shows the results of \ctbench using the methodology described in \cref{sec:implementation}. We further include the average and standard deviation obtained from independent runs in \cref{app:training_stability}, to validate the significance of our results. We find that \ctbench achieves consistent improvements in certified accuracy for almost all settings, accompanied by increases in natural accuracy in most cases. In particular, it establishes the new SOTA by a margin matching algorithmic advances everywhere except \cifar $\epsilon=\frac{8}{255}$, where we have $0.03\%$ lower certified accuracy compared to \citet{palma2024expressive} but $0.93\%$ higher natural accuracy. This proves the effectiveness of our implementation and the importance of setting batch norm statistics properly in certified training. We also observe the following:
(1) when $\epsilon$ is large, the claimed advantage of recent SOTA over \ibp drops significantly, \eg from $(100-93.10)/(100-93.62)-1=8.15\%$ relative certified error reduction to $(100-93.8)/(100-93.95)-1=2.48\%$ on \mnist $\epsilon=0.3$; (2) when the model has sufficient capacity, \eg, on \mnist $\epsilon=0.1$, certified training can get close to the natural accuracy of standard training (99.18\% for \mtlibp vs 99.50\% for standard training), and they also get similar adversarial accuracy to adversarial training (98.58\% for \taps vs 98.95\% for \edac), while certified accuracy is boosted (98.52\% for \taps vs almost 0\% for \edac); (3) when $\epsilon$ is large, certified training even gets better adversarial accuracy than \pgd training (94.66\% for \taps vs 93.83\% for \pgd on \mnist $\epsilon=0.3$ and 36.11\% for \sabr vs 35.93\% for \pgd on \cifar $\epsilon=\frac{8}{255}$), but there is still a gap between the adversarial accuracy of the SOTA adversarial training methods and that of the SOTA certified training methods, as well as a similar gap for natural accuracy. We further include a comparison on another architecture, \cnnf, between \ctbench and the implementation of \citet{palma2024expressive} in \cref{app:expressive_comparison}, to validate the stability of \ctbench results across architectures.

\section{Evaluating and Understanding Certified Models} \label{sec:insight}
We now preform an extensive evaluation on models trained with \ctbench. Our evaluation provides insights into the current state of certified training and addresses several key questions, including the loss fragmentation (\cref{sec:fragmentation}), shared mistakes (\cref{sec:correlation}), model utilization (\cref{sec:utilization}), regularization strength (\cref{sec:prop_tightness}), and out-of-distribution generalization (\cref{sec:ood}).

\subsection{Loss Fragmentation} \label{sec:fragmentation}

\begin{figure*}
    \centering
    \includegraphics[width=.6\linewidth]{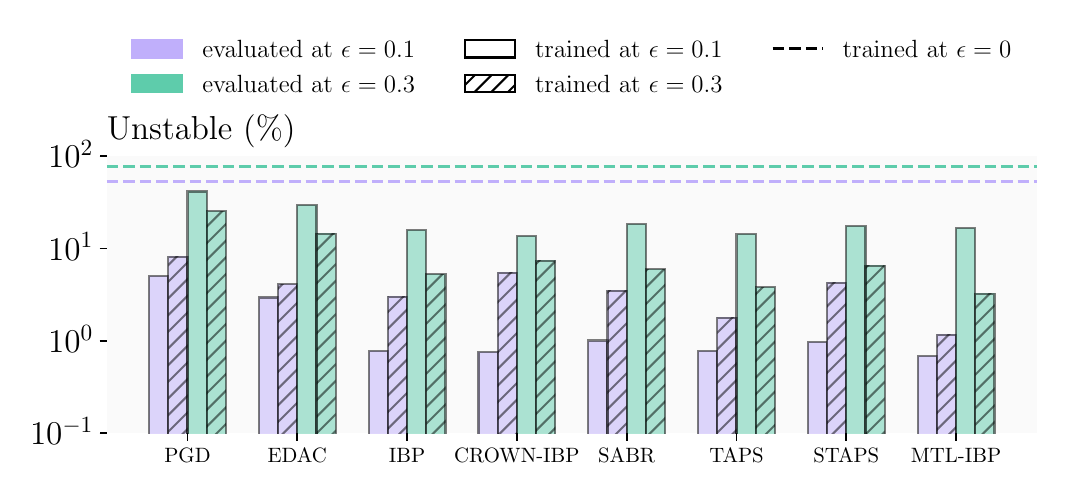}
    \caption{Ratio of unstable neurons for models trained on \mnist with different methods and $\epsilon$.} \label{fig:fragmentation_mnist}
\end{figure*}

ReLU networks are known to have a fragmented loss surface over the input space, due to the activation switch of neurons. Fragmentation leads to a non-smooth loss surface and increases the difficulty of finding a good approximation of the worst-case loss via gradient-based methods like \pgd. Further, SOTA complete certification algorithms relies on branching on different linear regions, and reducing the number of linear regions reduces the certification difficulty. Due to these reasons, in this section, we investigate the fragmentation of loss surfaces in certified models. Specifically, we answer: (1) do certified models have less fragmentation, thus easing adversarial search, and (2) how does the fragmentation change w.r.t. $\epsilon$?

Fragmentation is closely related to the number of unstable neurons, \ie, neurons that switch activation status in the neighborhood, as all fragments are defined by a group of unstable neurons. Vice versa, in most cases, a switching neuron introduces at least one fragmentation since every activation pattern defines a local linear function. Therefore, we can quantify the fragmentation by the ratio of unstable neurons. Since the exact ratio is NP-complete to compute, we use a heuristic but effective method to estimate it: first, a group of inputs is sampled in the input box; second, these inputs are fed into the model to get the corresponding activation pattern; finally, we count the ratio of unstable neurons observed in the sampled activations. This method always establishes a lower bound of the true ratio and gets arbitrarily close when sample size is large enough. In our experiments, we sample the noise 50 times from a standard Gaussian clipped to $[-1,1]$ and rescale it by $\epsilon$. This sampling focuses more on the neighborhood of the clean input and the boundary of the input box, where new fragments appear most likely. We find this sampling process extremely effective; empirically the ratio of unstable neurons observed is very close to the upper bounds derived by \ibp for certified models.

\cref{fig:fragmentation_mnist} shows the result of certified models trained at $\epsilon=0.1$ and $\epsilon=0.3$ on \mnist, respectively. We evaluate the fragmentation of every model at both $\epsilon=0.1$ and $\epsilon=0.3$.  First, we observe that both adversarial training and certified training greatly reduce loss fragmentation compared to standard training, even though many certified training algorithms involve no adversarial attacks. Second, comparing different training methods within each group of \purplebar and \greenbarhatched, we observe that certified training consistently has significantly less fragmentation than adversarial training when evaluated at the train radius, \eg, \ibp reduces fragmentation by 3x compared to \edac when trained and evaluated at $\epsilon=0.1$, facilitating the approximation of the worst case loss via adversarial attacks. This is consistent with the practice where a weak single-step attack is adopted in certified training \citep{palma2024expressive}, resulting in similar performance as strong attacks but improved efficiency. Third, comparing models trained at different $\epsilon$ (\purplebar vs \purplebarhatched and \greenbar vs \greenbarhatched), we observe that further increasing training $\epsilon$ does not necessarily reduce fragmentation, yet the trend is consistent with adversarial training. These observations prove that certified training can further boost the fragmentation reduction effect of adversarial training, thus introducing more local smoothness into the model. Consistent results on \cifar are included in \cref{app:additional_cifar} as \cref{fig:fragmentation_cifar}.

\subsection{Shared Mistakes} \label{sec:correlation}

\begin{table}
    \centering
    \caption{Observed count of common mistakes of models on \mnist against their expected values assuming independence across model mistakes.} \label{tb:common_mistake_mnist}
    \begin{adjustbox}{width=.95\linewidth,center}
    \begin{tabular}{ccccccccc}
        \toprule
        && \multicolumn{7}{c}{\# models succeeded} \\
        && 0 & 1 & 2 & 3 & 4 & 5 & 6 \\
        \cmidrule{3-9}
        \multirow{2}*{$\epsilon=0.1$} & obs. & 93 & 25 & 21 & 30 & 32 & 56 & 9743 \\
        & exp. & 0 & 0 & 0 & 1 & 37 & 900 & 9062 \\
        \midrule
        \multirow{2}*{$\epsilon=0.3$} & obs. & 452 & 73 & 53 & 51 & 80 & 111 & 9180 \\
        & exp. & 0 & 0 & 2 & 39 & 445 & 2698 & 6816 \\
        \bottomrule
    \end{tabular} 
    \end{adjustbox}
\end{table}

We now study the correlation between certified models, specifically: do certified models share common mistakes?

We consider models trained by six certified training methods on \mnist at $\epsilon=0.1$ and $\epsilon=0.3$ and calculate the distribution of their common mistakes. Specifically, we count the number of models that successfully certify the input, for each sample in the test set containing 10k samples. The observed value is compared with the expected value, defined as expected counts when models with the same certified accuracy make mistakes independently (rounded to the closest integer if necessary), \eg two models with 90\% certified accuracy are expected to have 81\% of inputs certified by both. The result is shown in \cref{tb:common_mistake_mnist}. Accordingly, certified models share many mistakes, as the number of samples that are certified by none of these models systematically exceeds the expected value. In addition, the number of samples that are certified by all six models is much larger than the corresponding expected value as well. These facts suggest that there could be an intrinsic difficulty score for each input, thus curriculum learning \citep{bengio09curriculum,IonescuALPPF16} could be a promising direction to improve certified training. This phenomenon is also observed across different certification methods, different model architectures and different datasets, as shown in \cref{app:correlation}, confirming that this is not a context-specific property but rather an intrinsic property of certified models.

\subsection{Model Utilization} \label{sec:utilization}

Model utilization represents how much the model capacity is utilized for the task. \ibp is shown to systematically deactivate neurons \citep{ShiWZYH21} to gain precision. However, it is not yet clear whether more advanced certified training methods deactivate fewer neurons, thus utilizing the model capacity better.

\begin{figure*}
    \centering
    \includegraphics[width=.6\linewidth]{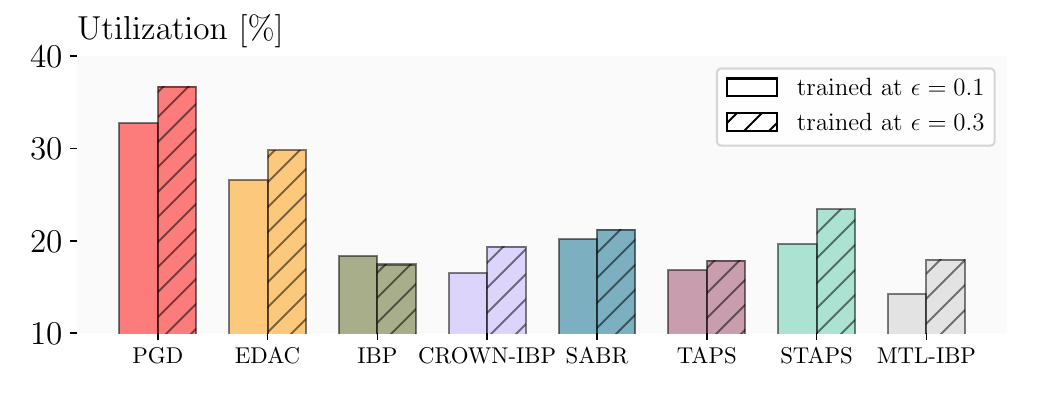}
    \caption{Model utilization for models trained on \mnist with different methods and $\epsilon$. We note that standard training has 42.99\% utilization.} \label{fig:utilization_mnist}
\end{figure*}

We define model utilization to be the ratio of neurons activated by the clean input. \cref{fig:utilization_mnist} visualizes the result for models trained on \mnist at $\epsilon=0.1$ and $\epsilon=0.3$. Surprisingly, we find that more advanced certified training methods, \taps and \mtlibp, deactivate more neurons than \ibp on \mnist $\epsilon=0.1$. This is previously believed to be detrimental \citep{ShiWZYH21}, but these models achieve better natural and certified accuracy than \ibp. Interestingly, these methods, but not \ibp, can retain more utilization on $\epsilon=0.3$ for better performance where \ibp struggles to keep high natural accuracy. Further, we observe that the advanced adversarial training method, \edac, shows similar behaviors to \taps and \mtlibp, and gets higher adversarial accuracy than \pgd. This suggests that the ability to adaptively keep necessary utilization could be crucial to both adversarial and certified robustness. Since dying neurons \citep{lu2019dying} are hard to activate again, future work on better warmup \citep{ShiWZYH21} could be beneficial, as  \ibp still struggles to keep necessary model utilization even with the improvements by \citet{ShiWZYH21}. More results on \cifar are included in \cref{app:additional_cifar} as \cref{fig:utilization_cifar}. There, however, all certified training methods cannot activate more neurons when $\epsilon$ is large, just like \ibp in \mnist, while adversarial training methods show similar behavior to \mnist. This suggests that the current certified training methods have not fully solved the utilization problem, especially when $\epsilon$ is large.

\subsection{Regularization Strength} \label{sec:prop_tightness}

Previous work \citep{mao2024understanding} has shown that \ibp bounds are close to optimal bounds for \ibp-based certified training, and this condition is established via strong constraints on the model parameters. They quantify this regularization effect by \emph{propagation tightness}, defined to be the ratio between the optimal bound radius and the \ibp bound radius, approximating the ReLU network locally with a linear replacement. Intuitively, a close-to-1 propagation tightness means \ibp bounds approximately match the exact bounds, and a close-to-0 propagation tightness means \ibp bounds are far from the exact bounds. In addition, a high propagation tightness implies strong regularization itself. In this section, we extend the study of propagation tightness to more certified training methods and investigate how it interacts with certified accuracy.
Specifically, using propagation tightness as the representative of regularization strength, we answer: (1) does less regularization lead to better certified models, and (2) how does the input radius $\epsilon$ affect this?

\begin{figure*}[t]
    \centering
    \begin{subfigure}[t]{.2329\linewidth} %
        \centering
        \includegraphics[width=.93\linewidth]{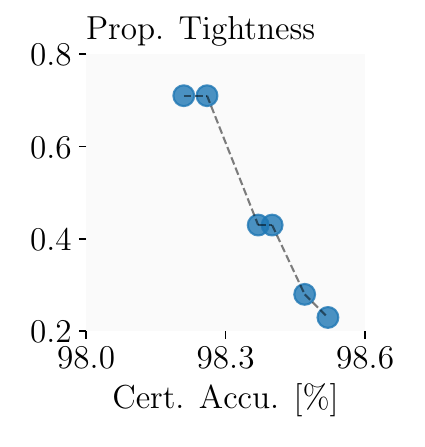}
        \caption{\mnist \\ \;\;\; $\epsilon=0.1$}
        \label{fig:tightness0.1}
    \end{subfigure}
    \begin{subfigure}[t]{.2381\linewidth} %
        \centering
        \includegraphics[width=.93\linewidth]{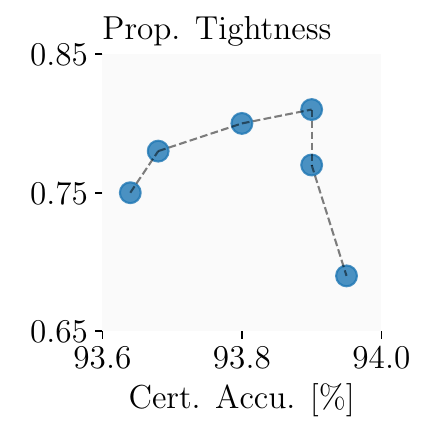}
        \caption{\mnist \\ \;\;\;\; $\epsilon=0.3$}
        \label{fig:tightness0.3}
    \end{subfigure}
    \begin{subfigure}[t]{.2225\linewidth} %
        \centering
        \includegraphics[width=.93\linewidth]{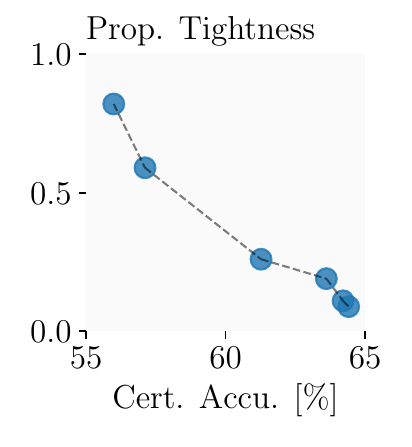}
        \caption{\cifar \\ \; $\epsilon=\frac{2}{255}$}
        \label{fig:tightness2.255}
    \end{subfigure}
    \begin{subfigure}[t]{.2381\linewidth} %
        \centering
        \includegraphics[width=.93\linewidth]{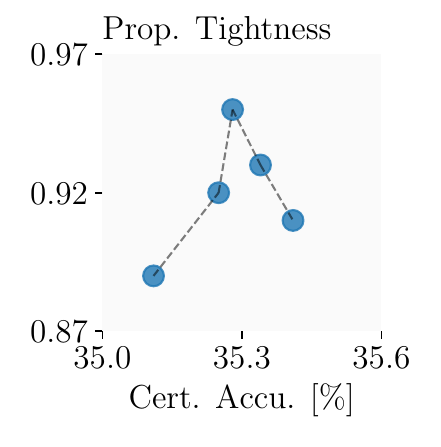}
        \caption{\cifar \\ \;\; $\epsilon=\frac{8}{255}$}
        \label{fig:tightness8.255}
    \end{subfigure}
    \caption{Certified accuracy vs. propagation tightness for models trained on \mnist and \cifar.} \label{fig:tightness}
\end{figure*}

\cref{fig:tightness} shows the interaction between certified accuracy and propagation tightness for certified models trained on \mnist and \cifar. When $\epsilon$ is small (\cref{fig:tightness0.1} and \cref{fig:tightness2.255}), certified accuracy has a perfectly negative correlation with propagation tightness, \ie, better certified models exhibits less regularization consistently; the best models have close-to-0 propagation tightness. However, when $\epsilon$ is large (\cref{fig:tightness0.3} and \cref{fig:tightness8.255}), the correlation is not as clear, and the best model in certified accuracy does not necessarily have the lowest propagation tightness. Instead, models with similar propagation tightness can have significantly different certified accuracy.  Nevertheless, models trained for larger radius exhibits much higher propagation tightness. Therefore, we conclude that reducing regularization strength cleverly is crucial for certified training, and the effect is more pronounced when $\epsilon$ is small, while improper reduction could hurt certified accuracy, especially when $\epsilon$ is large. This is consistent with the observation made in \citet{MuellerEFV22} and \citet{palma2024expressive} that the best models for small $\epsilon$ often require much less regularization.

\subsection{Out-of-Distribution Generalization} \label{sec:ood}

Out-of-distribution (OOD) generalization is closely related to adversarial robustness \citep{GilmerFCC19}. However, the interaction between certified robustness and OOD generalization is not yet clear. We thus investigate the OOD generalization of certified models and answer: (1) do certified models generalize to OOD data, and (2) how does this compare to adversarial training?

We use \mnist-C \citep{mnistcorrupted} to evaluate OOD generalization, defined to be the ratio between OOD accuracy and natural accuracy. \mnist-C includes 15 carefully chosen corruptions, covering a broad range of corruptions that are not covered by adversarial robustness while preserving the semantics. We evaluate models trained with both adversarial training and certified training under $\epsilon=0.1$ and $\epsilon=0.3$, and report the corresponding OOD accuracy of the model trained via standard training. We note that none of the models has seen these corruptions during training.

\begin{figure}
    \centering
    \adjustbox{minipage=[c]{\linewidth}}{

        \includegraphics[width=\linewidth]{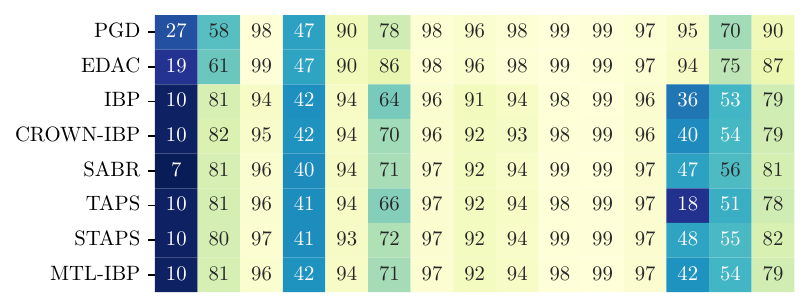}

        \includegraphics[width=\linewidth]{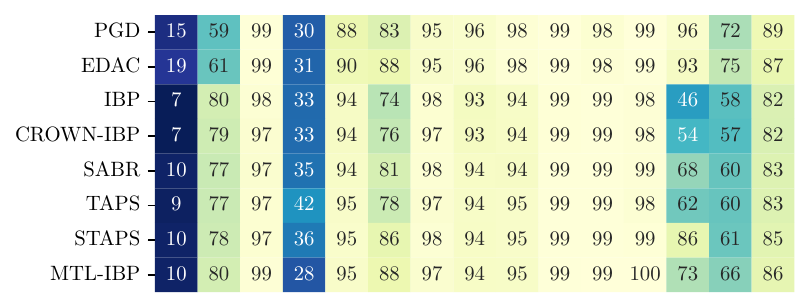}

        \hfill
        \includegraphics[width=.948\linewidth]{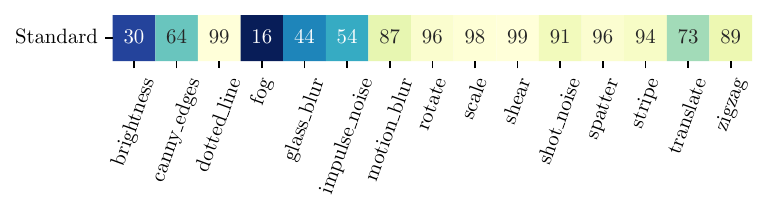}

    }
    \caption{Out-of-distribution generalization evaluated on \mnist-C for models trained on \mnist at $\epsilon=0.3$ (top), $\epsilon=0.1$ (middle) and standard training (bottom).} \label{fig:ood_mnist}
    \vspace{-5mm}
\end{figure}

\cref{fig:ood_mnist} depicts the result of OOD generalization for each model on all corruptions. We observe the following: (1) certified training improves OOD generalization compared to standard training except on the \emph{brightness} corruption where both adversarial and certified training fail; (2) certified training shows different OOD generalization patterns to adversarial training, \eg, certified training boosts generalization on the \emph{canny edges} corruption while adversarial training wins on the \emph{stripe} corruption. In general, we find that certified training either greatly boosts the OOD generalization or significantly downgrades the OOD generalization depending on the corruption, and the failure cases are usually those in which adversarial training performs worse than or similarly to standard training. Therefore, we hypothesize that these corruptions are at odds with adversarial robustness. Further, different training $\epsilon$ does not significantly affect the OOD generalization except few cases, and ranking in certified accuracy does not show strong correlations with the ranking in OOD generalization. Overall, these results suggest that certified training has the potential to improve OOD generalization to corruptions that standard training struggles with, and the effect is exaggerated when adversarial training improves over standard training. Consistent results on \cifar-C \citep{hendrycks2018benchmarking} are included in \cref{app:additional_cifar} as \cref{fig:ood_cifar}.

\section{Discussion} \label{sec:future}

\paragraph{Future Directions} \cref{sec:fragmentation} shows that certified training consistently reduces loss fragmentation, which also benefits adversarial attacks. Therefore, future works may explore architectures and training algorithms that explicitly have little loss fragmentation. In addition, \cref{sec:correlation} shows that certified models share mistakes on some hard samples, thus curriculum learning with some well-defined difficulty ranking could facilitate training, where optimization has been known to be particularly hard \citep{jovanovic2022paradox}. Moreover, \cref{sec:utilization} shows that even the most advanced certified training method, \mtlibp, struggles to keep necessary model utilization on large $\epsilon$. Further, \cref{sec:prop_tightness} finds that reducing regularization has a different consequence in small and large radius settings; while reducing regularization benefits small radius, it risks decreased performance in large radius settings. Overall, future work is still required to improve the learning process of certified training. Despite the challenges, \cref{sec:ood} finds that certified models can have surprising and qualitatively different improvement on OOD generalization, which could be a promising application for certified training beyond certified robustness. 

\paragraph{Limitations} We only consider deterministic certified training in \ctbench, while randomized certified robustness \citep{CohenRK19} has also made substantial progress. Moreover, we only consider the adversarial robustness, while other types of robustness, such as robustness against patch attacks \citep{SalmanJWM22} is also important. Finally, we only focus on $L_\infty$ robustness because there exists no \textit{scalable} deterministic certified training algorithm regarding other norms, and leave them as future work.

\paragraph{Connection to Randomized Certified Training} The issue of unfair comparison highlighted in this work may generalize to the domains of randomized certified robustness, particularly Randomized Smoothing (RS) as introduced by \citet{CohenRK19}. Recent RS literature shows considerable difference in evaluation regarding important aspects such as network architectures, training procedures, hyperparameter configurations, and the noise distributions employed for certification. These inconsistencies suggest that unfair comparisons may also be prevalent in RS studies. Therefore, we believe that a similar unified library and benchmark for randomized certified training would be beneficial to the community, allowing for fair comparisons and systematic hyperparameter tuning. We leave this as future work.

\section{Conclusion}
We introduced \ctbench, a unified library and high-quality benchmark for deterministic certified training on $L_\infty$-norm robustness. Based on \ctbench, we extensively evaluated certified models trained via state-of-the-art methods, analyzing their regularization strength and utilities. Our analysis reveals that certified training schemes can reduce loss fragmentation, adaptively keep model utilization, make shared mistakes, and generalize well on data with certain corruptions. We are confident that the insights and tools provided by \ctbench will facilitate future research on certified training and its applications.

\message{^^JLASTBODYPAGE \thepage^^J}

\section*{Reproducibility Statement}

We release the complete codebase of \ctbench, including the implementation of all certified training methods and the model checkpoints for the benchmark. The codebase is available at %
\href{https://github.com/eth-sri/CTBench}{https://github.com/eth-sri/CTBench}. 
A complete description of the experiment setup and hyperparameters is provided in \cref{app:exp_setting}.

\clearpage

\section*{Acknowledgements}

This work has been done as part of the EU grant ELSA (European Lighthouse on Secure and Safe AI, grant agreement no. 101070617) and the SERI grant SAFEAI (Certified Safe, Fair and Robust Artificial Intelligence, contract no. MB22.00088). Views and opinions expressed are however those of the authors only and do not necessarily reflect those of the European Union or European Commission. Neither the European Union nor the European Commission can be held responsible for them. 

This research was partially funded by the Ministry of Education and Science of Bulgaria (support for INSAIT, part of the Bulgarian National Roadmap for Research Infrastructure).

The work has received funding from the Swiss State Secretariat for Education, Research and Innovation (SERI).

\section*{Impact Statement}
This paper presents work whose goal is to advance the field of Machine Learning. There are many potential societal consequences of our work, none of which we feel must be specifically highlighted here.

\bibliography{references}

\begin{thebibliography}{55}
\providecommand{\natexlab}[1]{#1}
\providecommand{\url}[1]{\texttt{#1}}
\expandafter\ifx\csname urlstyle\endcsname\relax
  \providecommand{\doi}[1]{doi: #1}\else
  \providecommand{\doi}{doi: \begingroup \urlstyle{rm}\Url}\fi

\bibitem[Athalye et~al.(2018)Athalye, Carlini, and Wagner]{AthalyeC018}
Athalye, A., Carlini, N., and Wagner, D.~A.
\newblock Obfuscated gradients give a false sense of security: Circumventing
  defenses to adversarial examples.
\newblock In \emph{Proc. of ICML}, 2018.

\bibitem[Baader et~al.(2024)Baader, Mueller, Mao, and
  Vechev]{baader2024expressivity}
Baader, M., Mueller, M.~N., Mao, Y., and Vechev, M.
\newblock Expressivity of re{LU}-networks under convex relaxations.
\newblock In \emph{Proc. ICLR}, 2024.

\bibitem[Balauca et~al.(2024)Balauca, Müller, Mao, Baader, Fischer, and
  Vechev]{balauca2024overcoming}
Balauca, S., Müller, M.~N., Mao, Y., Baader, M., Fischer, M., and Vechev, M.
\newblock Overcoming the paradox of certified training with gaussian smoothing,
  2024.

\bibitem[Balunovic \& Vechev(2020)Balunovic and Vechev]{BalunovicV20}
Balunovic, M. and Vechev, M.~T.
\newblock Adversarial training and provable defenses: Bridging the gap.
\newblock In \emph{Proc. of ICLR}, 2020.

\bibitem[Bengio et~al.(2009)Bengio, Louradour, Collobert, and
  Weston]{bengio09curriculum}
Bengio, Y., Louradour, J., Collobert, R., and Weston, J.
\newblock Curriculum learning.
\newblock In \emph{Proc. of ICML}, 2009.

\bibitem[Bunel et~al.(2020)Bunel, Lu, Turkaslan, Torr, Kohli, and
  Kumar]{BunelLTTKK20}
Bunel, R., Lu, J., Turkaslan, I., Torr, P. H.~S., Kohli, P., and Kumar, M.~P.
\newblock Branch and bound for piecewise linear neural network verification.
\newblock \emph{J. Mach. Learn. Res.}, 2020.

\bibitem[Cohen et~al.(2019)Cohen, Rosenfeld, and Kolter]{CohenRK19}
Cohen, J.~M., Rosenfeld, E., and Kolter, J.~Z.
\newblock Certified adversarial robustness via randomized smoothing.
\newblock In \emph{Proc. of ICML}, 2019.

\bibitem[Croce \& Hein(2020)Croce and Hein]{Croce020a}
Croce, F. and Hein, M.
\newblock Reliable evaluation of adversarial robustness with an ensemble of
  diverse parameter-free attacks.
\newblock In \emph{Proc. of ICML}, 2020.

\bibitem[{De Palma} et~al.(2022){De Palma}, Bunel, Dvijotham, Kumar, and
  Stanforth]{PalmaIBPR22}
{De Palma}, A., Bunel, R., Dvijotham, K., Kumar, M.~P., and Stanforth, R.
\newblock {IBP} regularization for verified adversarial robustness via
  branch-and-bound, 2022.

\bibitem[{De Palma} et~al.(2024){De Palma}, Bunel, Dvijotham, Kumar, Stanforth,
  and Lomuscio]{palma2024expressive}
{De Palma}, A., Bunel, R.~R., Dvijotham, K.~D., Kumar, M.~P., Stanforth, R.,
  and Lomuscio, A.
\newblock Expressive losses for verified robustness via convex combinations.
\newblock In \emph{Proc. of ICLR}, 2024.

\bibitem[Ferrari et~al.(2022)Ferrari, M{\"{u}}ller, Jovanovic, and
  Vechev]{FerrariMJV22}
Ferrari, C., M{\"{u}}ller, M.~N., Jovanovic, N., and Vechev, M.~T.
\newblock Complete verification via multi-neuron relaxation guided
  branch-and-bound.
\newblock In \emph{Proc. of ICLR}, 2022.

\bibitem[Gehr et~al.(2018)Gehr, Mirman, Drachsler{-}Cohen, Tsankov, Chaudhuri,
  and Vechev]{GehrMDTCV18}
Gehr, T., Mirman, M., Drachsler{-}Cohen, D., Tsankov, P., Chaudhuri, S., and
  Vechev, M.~T.
\newblock {AI2:} safety and robustness certification of neural networks with
  abstract interpretation.
\newblock In \emph{Proc. of S\&P}, 2018.

\bibitem[Gilmer et~al.(2019)Gilmer, Ford, Carlini, and Cubuk]{GilmerFCC19}
Gilmer, J., Ford, N., Carlini, N., and Cubuk, E.~D.
\newblock Adversarial examples are a natural consequence of test error in
  noise.
\newblock In \emph{Proc. of ICML}, 2019.

\bibitem[Goodfellow et~al.(2015)Goodfellow, Shlens, and
  Szegedy]{GoodfellowSS14}
Goodfellow, I.~J., Shlens, J., and Szegedy, C.
\newblock Explaining and harnessing adversarial examples.
\newblock In \emph{Proc. of ICLR}, 2015.

\bibitem[Gowal et~al.(2018)Gowal, Dvijotham, Stanforth, Bunel, Qin, Uesato,
  Arandjelovic, Mann, and Kohli]{GowalIBP2018}
Gowal, S., Dvijotham, K., Stanforth, R., Bunel, R., Qin, C., Uesato, J.,
  Arandjelovic, R., Mann, T.~A., and Kohli, P.
\newblock On the effectiveness of interval bound propagation for training
  verifiably robust models.
\newblock 2018.

\bibitem[He et~al.(2015)He, Zhang, Ren, and Sun]{HeZRS15}
He, K., Zhang, X., Ren, S., and Sun, J.
\newblock Delving deep into rectifiers: Surpassing human-level performance on
  imagenet classification.
\newblock In \emph{Proc. of ICCV}, 2015.

\bibitem[Hendrycks \& Dietterich(2019)Hendrycks and
  Dietterich]{hendrycks2018benchmarking}
Hendrycks, D. and Dietterich, T.
\newblock Benchmarking neural network robustness to common corruptions and
  perturbations.
\newblock In \emph{Proc. of ICLR}, 2019.

\bibitem[Ioffe \& Szegedy(2015)Ioffe and Szegedy]{IoffeS15}
Ioffe, S. and Szegedy, C.
\newblock Batch normalization: Accelerating deep network training by reducing
  internal covariate shift.
\newblock In \emph{Proc. ICML}, 2015.

\bibitem[Ionescu et~al.(2016)Ionescu, Alexe, Leordeanu, Popescu, Papadopoulos,
  and Ferrari]{IonescuALPPF16}
Ionescu, R.~T., Alexe, B., Leordeanu, M., Popescu, M., Papadopoulos, D.~P., and
  Ferrari, V.
\newblock How hard can it be? estimating the difficulty of visual search in an
  image.
\newblock In \emph{Proc. of ICLR}, 2016.

\bibitem[Izmailov et~al.(2018)Izmailov, Podoprikhin, Garipov, Vetrov, and
  Wilson]{IzmailovPGVW18}
Izmailov, P., Podoprikhin, D., Garipov, T., Vetrov, D.~P., and Wilson, A.~G.
\newblock Averaging weights leads to wider optima and better generalization.
\newblock In \emph{Proc. of UAI}, 2018.

\bibitem[Jovanović et~al.(2022)Jovanović, Balunović, Baader, and
  Vechev]{jovanovic2022paradox}
Jovanović, N., Balunović, M., Baader, M., and Vechev, M.
\newblock On the paradox of certified training.
\newblock In \emph{Proc. of ICML}, 2022.

\bibitem[Katz et~al.(2017)Katz, Barrett, Dill, Julian, and
  Kochenderfer]{KatzBDJK17}
Katz, G., Barrett, C.~W., Dill, D.~L., Julian, K., and Kochenderfer, M.~J.
\newblock Reluplex: An efficient {SMT} solver for verifying deep neural
  networks.
\newblock In \emph{{CAV} {(1)}}, volume 10426 of \emph{Lecture Notes in
  Computer Science}, pp.\  97--117. Springer, 2017.

\bibitem[Kingma \& Ba(2015)Kingma and Ba]{KingmaB14}
Kingma, D.~P. and Ba, J.
\newblock Adam: {A} method for stochastic optimization.
\newblock In \emph{Proc. of ICLR}, 2015.

\bibitem[Krizhevsky et~al.(2009)Krizhevsky, Hinton,
  et~al.]{krizhevsky2009learning}
Krizhevsky, A., Hinton, G., et~al.
\newblock Learning multiple layers of features from tiny images.
\newblock 2009.

\bibitem[Le \& Yang(2015)Le and Yang]{Ya2015tinyimagenet}
Le, Y. and Yang, X.~S.
\newblock Tiny imagenet visual recognition challenge, 2015.

\bibitem[LeCun et~al.(2010)LeCun, Cortes, and Burges]{lecun2010mnist}
LeCun, Y., Cortes, C., and Burges, C.
\newblock Mnist handwritten digit database.
\newblock \emph{ATT Labs [Online]. Available:
  http://yann.lecun.com/exdb/mnist}, 2010.

\bibitem[Lee et~al.(2021)Lee, Lee, Park, and Lee]{LeeLPL21}
Lee, S., Lee, W., Park, J., and Lee, J.
\newblock Towards better understanding of training certifiably robust models
  against adversarial examples.
\newblock In \emph{Proc. NeurIPS}, 2021.

\bibitem[Li et~al.(2023)Li, Xie, and Li]{LiXL23}
Li, L., Xie, T., and Li, B.
\newblock Sok: Certified robustness for deep neural networks.
\newblock In \emph{{SP}}, pp.\  1289--1310. {IEEE}, 2023.

\bibitem[Lu et~al.(2019)Lu, Shin, Su, and Karniadakis]{lu2019dying}
Lu, L., Shin, Y., Su, Y., and Karniadakis, G.~E.
\newblock Dying relu and initialization: Theory and numerical examples, 2019.

\bibitem[Lyu et~al.(2024)Lyu, Shaikh, Shpilevskiy, Shelhamer, and
  Lécuyer]{lyu2024adaptiverandomizedsmoothingcertified}
Lyu, S., Shaikh, S., Shpilevskiy, F., Shelhamer, E., and Lécuyer, M.
\newblock Adaptive randomized smoothing: Certified adversarial robustness for
  multi-step defences, 2024.
\newblock URL \url{https://arxiv.org/abs/2406.10427}.

\bibitem[Madry et~al.(2018)Madry, Makelov, Schmidt, Tsipras, and
  Vladu]{MadryMSTV18}
Madry, A., Makelov, A., Schmidt, L., Tsipras, D., and Vladu, A.
\newblock Towards deep learning models resistant to adversarial attacks.
\newblock In \emph{Proc. of ICLR}, 2018.

\bibitem[Mao et~al.(2023)Mao, M{\"{u}}ller, Fischer, and Vechev]{MaoM0V23}
Mao, Y., M{\"{u}}ller, M.~N., Fischer, M., and Vechev, M.~T.
\newblock Connecting certified and adversarial training.
\newblock In \emph{Proc. of NeurIPS}, 2023.

\bibitem[Mao et~al.(2024)Mao, Mueller, Fischer, and
  Vechev]{mao2024understanding}
Mao, Y., Mueller, M.~N., Fischer, M., and Vechev, M.
\newblock Understanding certified training with interval bound propagation.
\newblock In \emph{Proc. of ICLR}, 2024.

\bibitem[Mao et~al.(2025)Mao, Zhang, and Vechev]{mao2025expressiveness}
Mao, Y., Zhang, Y., and Vechev, M.
\newblock On the expressiveness of multi-neuron convex relaxations, 2025.
\newblock URL \url{https://arxiv.org/abs/2410.06816}.

\bibitem[Mirman et~al.(2018)Mirman, Gehr, and Vechev]{MirmanGV18}
Mirman, M., Gehr, T., and Vechev, M.~T.
\newblock Differentiable abstract interpretation for provably robust neural
  networks.
\newblock In \emph{Proc. of ICML}, 2018.

\bibitem[Mu \& Gilmer(2019)Mu and Gilmer]{mnistcorrupted}
Mu, N. and Gilmer, J.
\newblock {MNIST-C:} {A} robustness benchmark for computer vision, 2019.

\bibitem[M{\"{u}}ller et~al.(2023)M{\"{u}}ller, Eckert, Fischer, and
  Vechev]{MuellerEFV22}
M{\"{u}}ller, M.~N., Eckert, F., Fischer, M., and Vechev, M.~T.
\newblock Certified training: Small boxes are all you need.
\newblock In \emph{Proc. of ICLR}, 2023.

\bibitem[Salman et~al.(2022)Salman, Jain, Wong, and Madry]{SalmanJWM22}
Salman, H., Jain, S., Wong, E., and Madry, A.
\newblock Certified patch robustness via smoothed vision transformers.
\newblock In \emph{Proc. of CVPR}, 2022.

\bibitem[Shi et~al.(2021)Shi, Wang, Zhang, Yi, and Hsieh]{ShiWZYH21}
Shi, Z., Wang, Y., Zhang, H., Yi, J., and Hsieh, C.
\newblock Fast certified robust training with short warmup.
\newblock In \emph{Proc. of NeurIPS}, 2021.

\bibitem[Singh et~al.(2018)Singh, Gehr, Mirman, P{\"{u}}schel, and
  Vechev]{SinghGMPV18}
Singh, G., Gehr, T., Mirman, M., P{\"{u}}schel, M., and Vechev, M.~T.
\newblock Fast and effective robustness certification.
\newblock In \emph{Proc. of NeurIPS}, 2018.

\bibitem[Singh et~al.(2019)Singh, Gehr, P{\"{u}}schel, and Vechev]{SinghGPV19}
Singh, G., Gehr, T., P{\"{u}}schel, M., and Vechev, M.~T.
\newblock An abstract domain for certifying neural networks.
\newblock In \emph{Proc. of POPL}, 2019.

\bibitem[Soletskyi \& Dalrymple(2024)Soletskyi and
  Dalrymple]{soletskyi2024global}
Soletskyi, R. and Dalrymple, D.
\newblock Training safe neural networks with global sdp bounds, 2024.
\newblock URL \url{https://arxiv.org/abs/2409.09687}.

\bibitem[Szegedy et~al.(2014)Szegedy, Zaremba, Sutskever, Bruna, Erhan,
  Goodfellow, and Fergus]{SzegedyZSBEGF13}
Szegedy, C., Zaremba, W., Sutskever, I., Bruna, J., Erhan, D., Goodfellow,
  I.~J., and Fergus, R.
\newblock Intriguing properties of neural networks.
\newblock In \emph{Proc. of ICLR}, 2014.

\bibitem[Tjeng et~al.(2019)Tjeng, Xiao, and Tedrake]{TjengXT19}
Tjeng, V., Xiao, K.~Y., and Tedrake, R.
\newblock Evaluating robustness of neural networks with mixed integer
  programming.
\newblock In \emph{Proc. of ICLR}, 2019.

\bibitem[Tram{\`{e}}r et~al.(2020)Tram{\`{e}}r, Carlini, Brendel, and
  Madry]{TramerCBM20}
Tram{\`{e}}r, F., Carlini, N., Brendel, W., and Madry, A.
\newblock On adaptive attacks to adversarial example defenses.
\newblock In \emph{Proc. of NeurIPS}, 2020.

\bibitem[Wang et~al.(2023)Wang, Jha, Krishnamurthy, and
  Dvijotham]{wang2023efficient}
Wang, Z., Jha, S., Krishnamurthy, and Dvijotham.
\newblock Efficient symbolic reasoning for neural-network verification, 2023.
\newblock URL \url{https://arxiv.org/abs/2303.13588}.

\bibitem[Wong \& Kolter(2018)Wong and Kolter]{WongK18}
Wong, E. and Kolter, J.~Z.
\newblock Provable defenses against adversarial examples via the convex outer
  adversarial polytope.
\newblock In \emph{Proc. of ICML}, 2018.

\bibitem[Wong et~al.(2018)Wong, Schmidt, Metzen, and Kolter]{WongSMK18}
Wong, E., Schmidt, F.~R., Metzen, J.~H., and Kolter, J.~Z.
\newblock Scaling provable adversarial defenses.
\newblock In \emph{Proc. of NeurIPS}, 2018.

\bibitem[Wu \& Johnson(2021)Wu and Johnson]{WuPreciseBN21}
Wu, Y. and Johnson, J.
\newblock Rethinking "batch" in batchnorm, 2021.

\bibitem[Xu et~al.(2020)Xu, Shi, Zhang, Wang, Chang, Huang, Kailkhura, Lin, and
  Hsieh]{XuS0WCHKLH20}
Xu, K., Shi, Z., Zhang, H., Wang, Y., Chang, K., Huang, M., Kailkhura, B., Lin,
  X., and Hsieh, C.
\newblock Automatic perturbation analysis for scalable certified robustness and
  beyond.
\newblock In \emph{Proc. of NeurIPS}, 2020.

\bibitem[Xu et~al.(2021)Xu, Zhang, Wang, Wang, Jana, Lin, and
  Hsieh]{XuZ0WJLH21}
Xu, K., Zhang, H., Wang, S., Wang, Y., Jana, S., Lin, X., and Hsieh, C.
\newblock Fast and complete: Enabling complete neural network verification with
  rapid and massively parallel incomplete verifiers.
\newblock In \emph{Proc. of ICLR}, 2021.

\bibitem[Zhang et~al.(2018)Zhang, Weng, Chen, Hsieh, and Daniel]{ZhangWCHD18}
Zhang, H., Weng, T., Chen, P., Hsieh, C., and Daniel, L.
\newblock Efficient neural network robustness certification with general
  activation functions.
\newblock In \emph{Proc. of NeurIPS}, 2018.

\bibitem[Zhang et~al.(2020)Zhang, Chen, Xiao, Gowal, Stanforth, Li, Boning, and
  Hsieh]{ZhangCXGSLBH20}
Zhang, H., Chen, H., Xiao, C., Gowal, S., Stanforth, R., Li, B., Boning, D.~S.,
  and Hsieh, C.
\newblock Towards stable and efficient training of verifiably robust neural
  networks.
\newblock In \emph{Proc. of ICLR}, 2020.

\bibitem[Zhang et~al.(2022)Zhang, Wang, Xu, Li, Li, Jana, Hsieh, and
  Kolter]{ZhangWXLLJ22}
Zhang, H., Wang, S., Xu, K., Li, L., Li, B., Jana, S., Hsieh, C., and Kolter,
  J.~Z.
\newblock General cutting planes for bound-propagation-based neural network
  verification, 2022.

\bibitem[Zhang et~al.(2023)Zhang, Backes, and Zhang]{zhang2023generating}
Zhang, M., Backes, M., and Zhang, X.
\newblock Generating less certain adversarial examples improves robust
  generalization, 2023.

\end{thebibliography}
\bibliographystyle{icml2025}

\message{^^JLASTREFERENCESPAGE \thepage^^J}

\newpage
\appendix
\onecolumn

\section{Improvement Decomposition} \label{app:decomposition}

Decomposition of the universal modifications we made such as batch norm fixes and the hyperparameter tuning is not always possible, as these modifications allow additional vectors of hyperparameter for tuning. For example, we fix batch norm statistics in one batch rather than reset it multiple times as done in some original implementations, allowing $w_{\text{rob}}$ to be tuned within $[0,1]$, while in the literature $w_{\text{rob}}$ has to be fixed to 1. Therefore, we cannot formally decompose the effects of hyperparameter tuning and batch norm behaviors, as they are closely dependent on each other.

The literature results are run with three different random seeds, and only the best results among them are reported. This prevents us from substituting our fine-tuned hyperparameter to the original implementation because merely using the same hyperparameters even based on the original implementation hardly reproduces the same number as reported in the literature. In contrast, we run every experiment with the same fixed random seed to allow fair and faithful comparison. Nevertheless, we can showcase the effect for one setting: IBP on MNIST $\epsilon=0.3$. The literature reports 93.1\% certified accuracy, while the same hyperparameter results in 93.18\% in our implementation. Further tuning the hyperparameters as in the CTBench benchmark gets 93.8\%. While this proves the effectiveness of both the implementation and our hyperparameter tuning, we would like to note that based on previous arguments, this does not faithfully decompose the effect of hyperparameter tuning and batch norm changes, and such decomposition efforts are doomed to fail.

While full disentanglement is infeasible, we conduct preliminary studies to separate implementation advantages from hyperparameter tuning. \cref{tab:expressive_comparison} compares CNN5 performance using CTBench and the SOTA codebase, applying CNN7-tuned hyperparameters to both to reduce tuning bias, showing CTBench's universal implementation benefits. Additionally, \cref{tab:l1_effects} ($L_1$ regularization on IBP-trained networks) and \cref{fig:rob_accu_curves} (effects of varying $\lambda$ for SABR and STAPS and $\alpha$ for MTL-IBP on the robustness-accuracy trade-off) illustrate hyperparameter impact.

\section{Additional Discussions} \label{app:additional_discussion}

\subsection{Ablation on $L_1$ Regularization} \label{app:l1_effects}

\cref{tab:l1_effects} shows the effect of $L_1$ regularization on the certified accuracy of IBP-trained networks. We observe that $L_1$ regularization tuned within a small range of hyperparameter choices can improve certified accuracy in most cases, especially for small perturbation sizes.

\begin{table}[h]
    \caption{Effect of $L_1$ regularization for IBP-trained networks on different datasets and perturbation sizes.}
    \label{tab:l1_effects}
    \centering
    \resizebox{0.3\linewidth}{!}{
        \begin{tabular}{@{}cccc@{}}
            \toprule
            Setting & $L_1$ weight     & Nat. [\%] & Cert. [\%]     \\
            \midrule
            \multirow{4}{*}{MNIST 0.1}
                    & 0                & 98.92     & 98.21          \\
                    & $1\cdot 10^{-6}$ & 98.84     & 98.22          \\
                    & $2\cdot 10^{-6}$ & 98.87     & \textbf{98.26} \\
                    & $5\cdot 10^{-6}$ & 98.85     & 98.13          \\
            \midrule
            \multirow{4}{*}{MNIST 0.3}
                    & 0                & 98.52     & 93.56          \\
                    & $1\cdot 10^{-6}$ & 98.54     & \textbf{93.82} \\
                    & $2\cdot 10^{-6}$ & 98.51     & 93.66          \\
                    & $5\cdot 10^{-6}$ & 98.40     & 93.76          \\
            \midrule
            \multirow{4}{*}{CIFAR 2/255}
                    & 0                & 67.81     & 55.45          \\
                    & $1\cdot 10^{-6}$ & 67.49     & \textbf{55.99} \\
                    & $2\cdot 10^{-6}$ & 66.15     & 55.01          \\
                    & $5\cdot 10^{-6}$ & 65.41     & 54.11          \\
            \midrule
            \multirow{2}{*}{CIFAR 8/255}
                    & 0                & 48.51     & \textbf{35.28} \\
                    & $1\cdot 10^{-6}$ & 48.31     & 34.36          \\
            \midrule
            \multirow{4}{*}{TIN 1/255}
                    & 0                & 25.68     & 19.04          \\
                    & $2\cdot 10^{-6}$ & 26.26     & \textbf{19.82} \\
                    & $5\cdot 10^{-6}$ & 26.37     & 19.80          \\
                    & $1\cdot 10^{-5}$ & 26.77     & \textbf{19.82} \\
            \bottomrule
        \end{tabular}
    }
\end{table}

\subsection{Robustness-Accuracy Trade-off} \label{app:tradeoff}

\begin{figure}[htbp]
    \centering
    \vspace{-3mm}
    \includegraphics[width=0.4\linewidth]{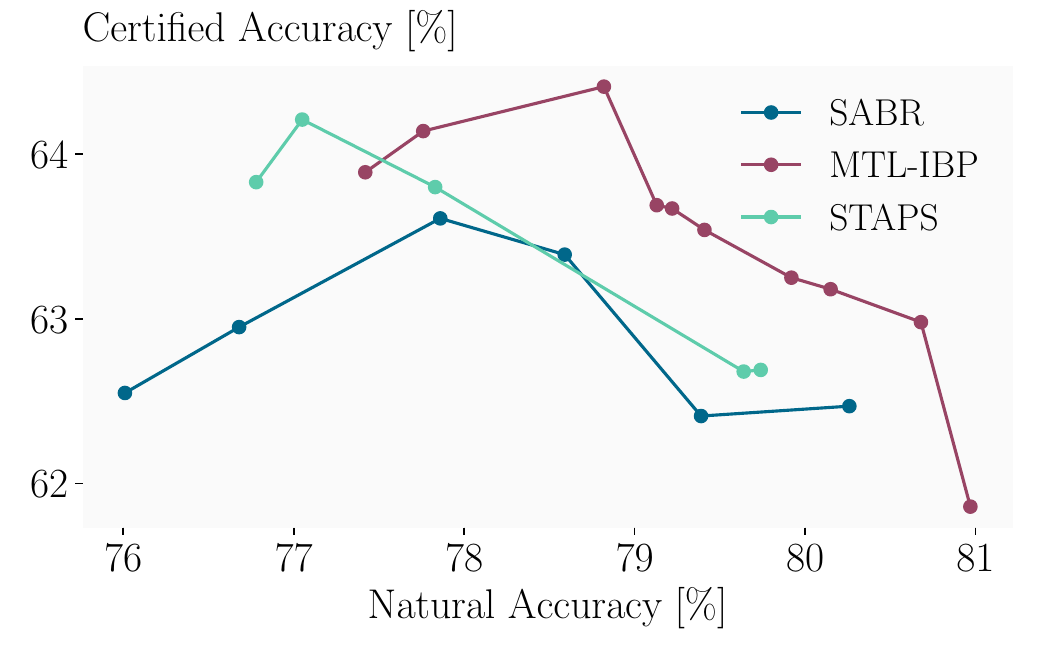}
    \vspace{-3mm}
    \caption{Robustness-Accuracy Trade-off for SABR, STAPS and MTL-IBP on CIFAR-10, $\epsilon=2/255$.}
    \label{fig:rob_accu_curves}
\end{figure}

The robustness-accuracy trade-off is well-known, where higher certified accuracy often comes at the cost of natural accuracy. Most methods, including SABR and MTL-IBP, have hyperparameters (e.g., $\lambda$, $\alpha$) that directly regulate this trade-off. Our goal, like in prior work, is to maximize certified accuracy, with natural accuracy improvements seen as a bonus. For completeness, we further provide robustness-accuracy curves, as shown in \cref{fig:rob_accu_curves}.

Consistent with prior work (\citet[Figure 7]{MuellerEFV22} and \citet[Figure 1]{palma2024expressive}), we observe that reducing regularization initially improves both robustness and natural accuracy, but beyond an optimal point, further reduction severely hurts certifiability while increasing natural accuracy.

\begin{figure}
    \centering
    \begin{subfigure}[t]{0.4\textwidth}
        \centering
        \includegraphics[width=\linewidth]{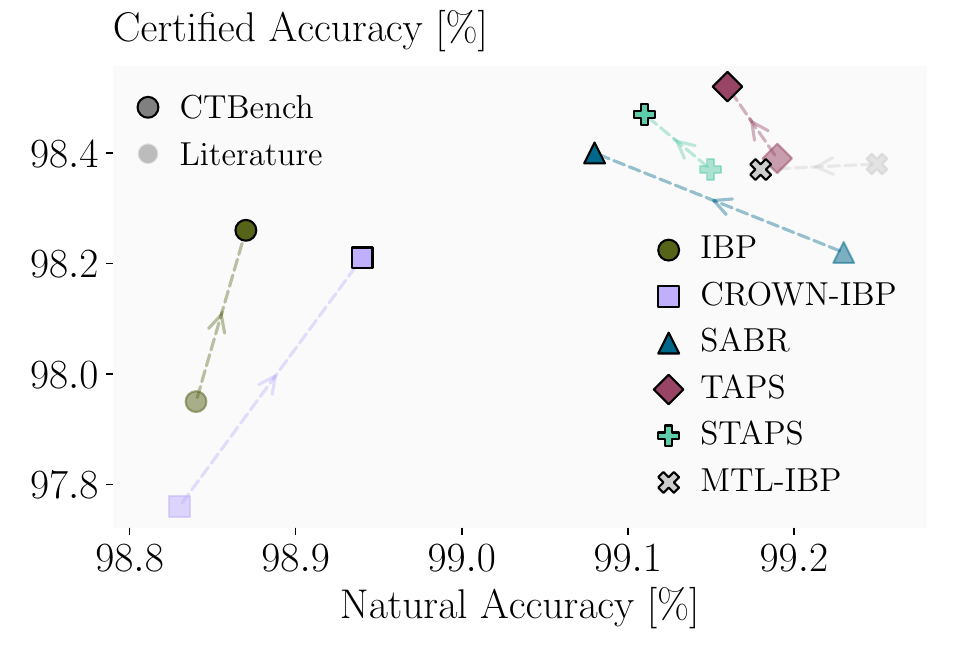}
        \vspace{-2mm}
        \caption{MNIST, $\epsilon=0.1$}
        \label{fig:sub1}
    \end{subfigure}
    \hspace{0.04\textwidth}
    \begin{subfigure}[t]{0.4\textwidth}
        \centering
        \includegraphics[width=\linewidth]{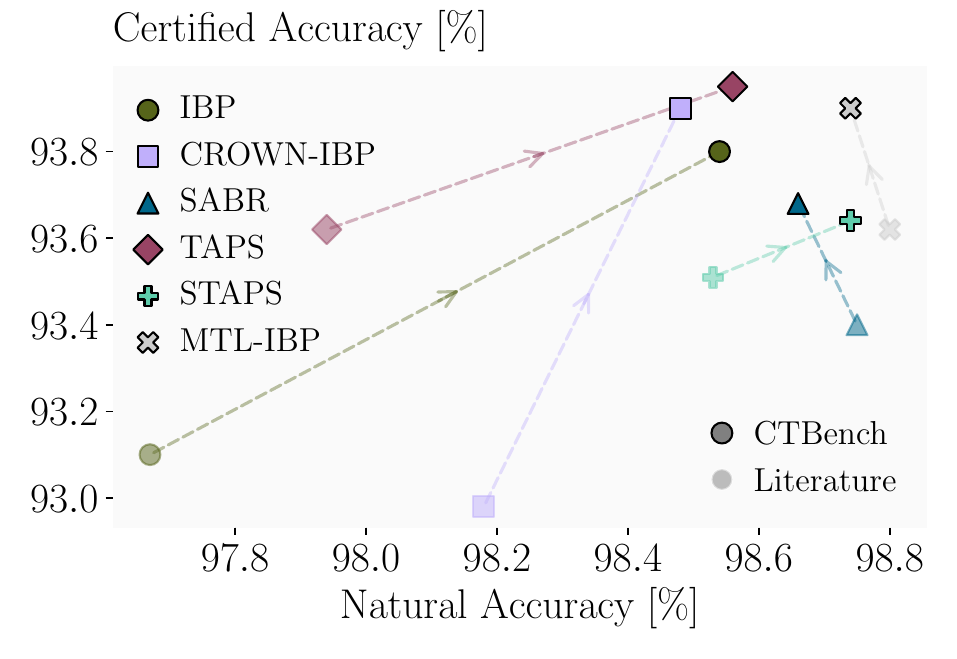}
        \vspace{-2mm}
        \caption{MNIST, $\epsilon=0.3$}
        \label{fig:sub2}
    \end{subfigure}

    \vspace{1em}

    \begin{subfigure}[t]{0.4\textwidth}
        \centering
        \includegraphics[width=\linewidth]{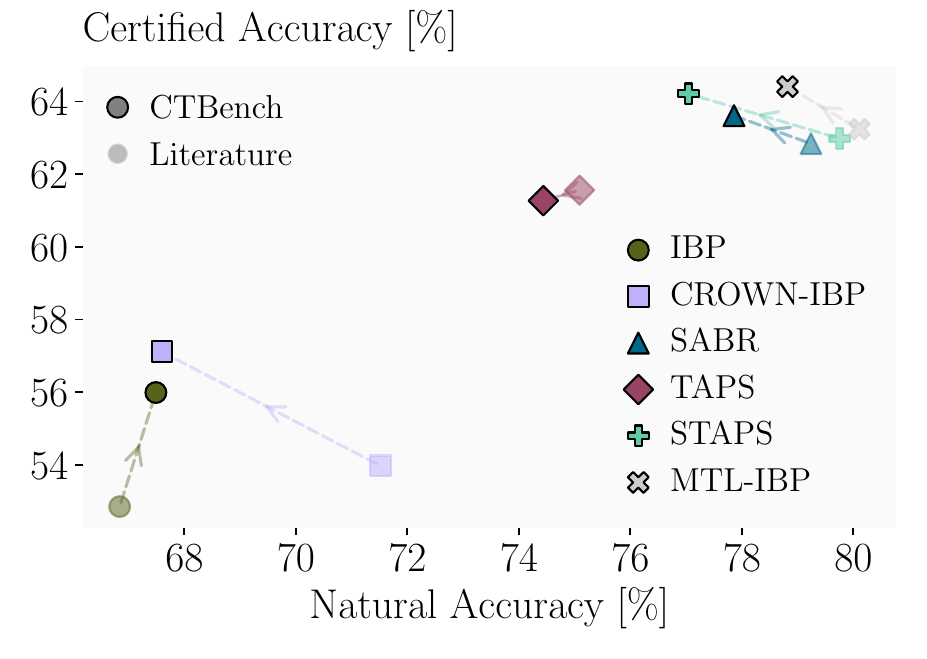}
        \vspace{-2mm}
        \caption{CIFAR-10, $\epsilon=2/255$}
        \label{fig:sub3}
    \end{subfigure}
    \hspace{0.04\textwidth}
    \begin{subfigure}[t]{0.4\textwidth}
        \centering
        \includegraphics[width=\linewidth]{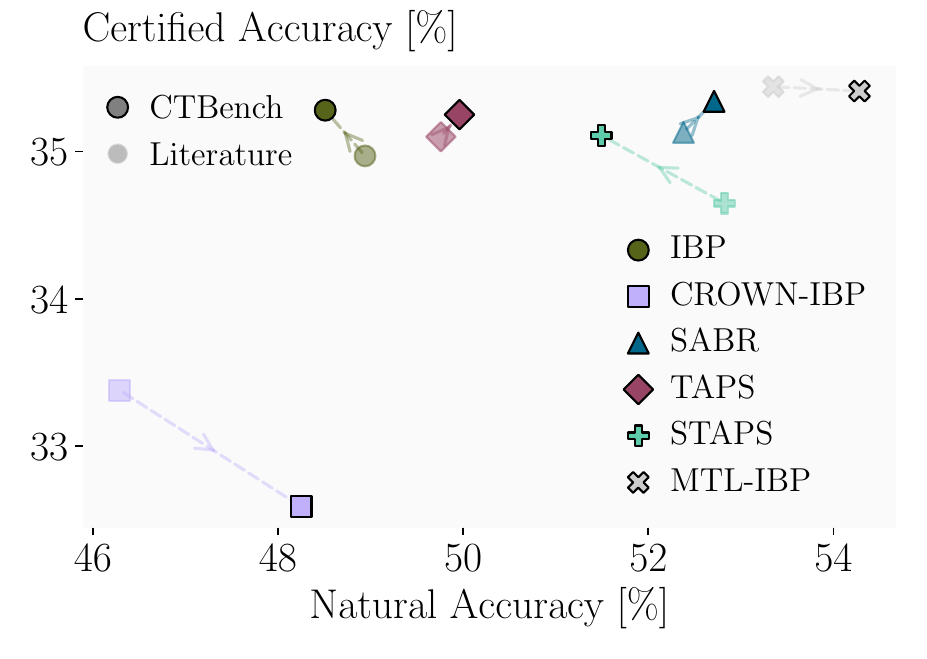}
        \vspace{-2mm}
        \caption{CIFAR-10, $\epsilon=8/255$}
        \label{fig:sub4}
    \end{subfigure}

    \vspace{1em}

    \begin{subfigure}[t]{0.4\textwidth}
        \centering
        \includegraphics[width=\linewidth]{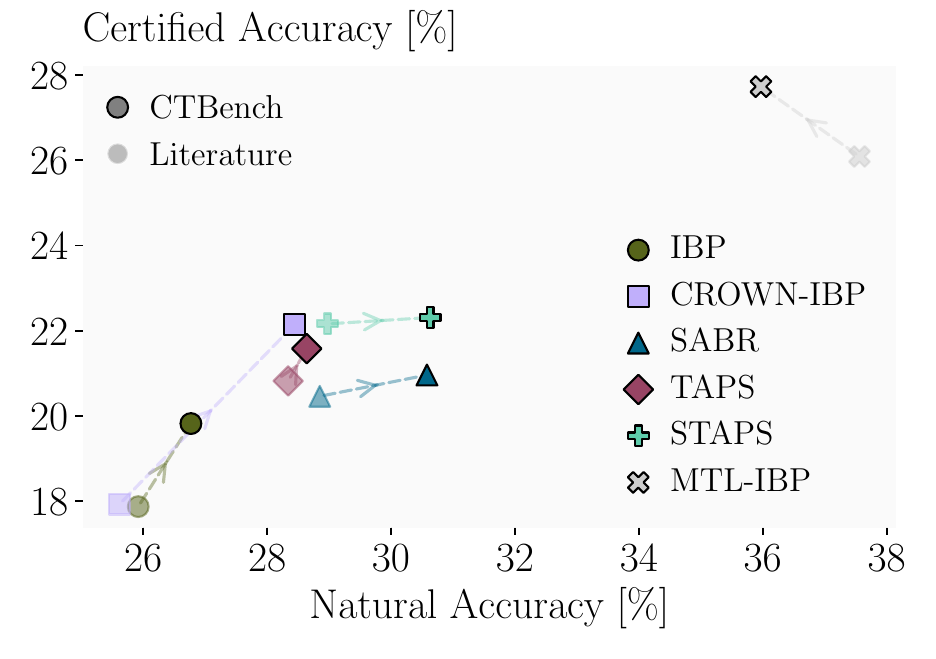}
        \vspace{-2mm}
        \caption{\TIN, $\epsilon=1/255$}
        \label{fig:sub5}
    \end{subfigure}

    \caption{Visualization of \ctbench improvements on the robustness-accuracy trade-off for each certification setting.
    }
    \label{fig:rob_accu_main}
    \vspace{-5mm}
\end{figure}

Additionally, we provide a visualization of how the robustness-accuracy trade-off changes for all certification settings in \cref{fig:rob_accu_main}, comparing \ctbench results with literature results. The transparent points represent previously reported best literature results, while the solid points represent \ctbench results.

\subsection{Extending $L_\infty$ Deterministic Certified Robustness to Other Norms} \label{app:other_norms}

Our work focuses on deterministic certified training using bound propagation for the $L_\infty$ norm, as it remains the most reliable and widely adopted approach for robustness guarantees. While \citet{LiXL23} explores various norms for certification, it also limits deterministic certified training to $L_\infty$, reflecting the current state of the field, with practical deterministic methods focused on $L_\infty$.

Certification under other norms, such as $L_2$, faces scalability challenges. For example, \citet{wang2023efficient} evaluate $L_2$ certification on small models with only 192 hidden nodes, while our CNN7 network has over 10M parameters, making their method impractical. Similarly, \citet{soletskyi2024global} use expensive SDP methods, limiting their approach to synthetic toy datasets (Spheres) and does not naturally extend to $L_\infty$. Furthermore, while \citet{soletskyi2024global} address $L_2$-norm robustness, their methods do not naturally extend to $L_\infty$.

Bound propagation is difficult for norms other than $L_\infty$. For example, for the $L_2$ norm, it is difficult to track the exact shape of the $L_2$ ball after passing through a linear and ReLU layer. To apply bound propagation, the typical $L_2$ ball with radius $\epsilon=1$ used in randomized smoothing must be over-approximated by the full $[0,1]^d$ input set, which guarantees meaningless results. Notably, \citet{wang2023efficient} do not attempt bound propagation. Developing new deterministic certification methods for other norms is out of the scope of this work.

Exploring deterministic certified training for other norms is a valuable future direction. However, due to scalability limitations and the lack of effective methods for other norms (even on MNIST), our focus remains on $L_\infty$.

\subsection{Comparison with $L_\infty$ Randomized Certified Robustness} \label{app:rand_smoothing}

In this section, we conduct a preliminary study on comparing $L_\infty$-norm robustness certified by RS to our results based on deterministic algorithms. Specifically, we compare the numbers by the state-of-the-art $L_\infty$-norm RS algorithm of \citet{lyu2024adaptiverandomizedsmoothingcertified} on CIFAR-10 at $\epsilon=2/255$ and $\epsilon=8/255$ with CTBench results in \cref{tab:rs_comparison}. We find that the current RS approaches yield lower certified accuracy compared to CTBench, in agreement with the literature where deterministic methods dominate the $L_\infty$-norm robustness.

\begin{table}[ht]
    \centering
    \caption{Comparison between Deterministic Certified Training (this work) and Adaptive Randomised Smoothing \citep{lyu2024adaptiverandomizedsmoothingcertified} on CIFAR-10.}
    \label{tab:rs_comparison}
    \resizebox{0.55\linewidth}{!}{
        \begin{tabular}{lccc}
            \toprule
            Method            & Nat. [\%] & Cert. at $\epsilon=\frac{2}{255}$ [\%] & Cert. at $\epsilon=\frac{8}{255}$ [\%] \\
            \midrule
            IBP               & 67.49    & 55.99           & /               \\
            IBP               & 48.51    & /               & 35.28           \\
            SABR              & 77.86    & 63.61           & /               \\
            SABR              & 52.71    & /               & 35.34           \\
            MTL-IBP           & 78.82    & \textbf{64.41}  & /               \\
            MTL-IBP           & 54.28    & /               & \textbf{35.41}  \\
            \midrule
            ARS $\sigma=0.12$ & 79.21    & 48.79           & 0.00            \\
            ARS $\sigma=0.25$ & 73.61    & 48.79           & 0.00            \\
            ARS $\sigma=0.50$ & 65.47    & \textbf{51.98}  & 8.96            \\
            ARS $\sigma=0.75$ & 57.71    & 44.64           & 11.91           \\
            ARS $\sigma=1.00$ & 49.96    & 40.71           & 14.07           \\
            ARS $\sigma=1.50$ & 39.86    & 31.41           & \textbf{14.57}  \\
            \bottomrule
        \end{tabular}
    }
\end{table}

\subsection{General Trends Across Datasets} \label{app:dataset_trends}

Across the datasets considered in this work, several performance trends emerge, offering insights into how different certification and training methods generalize. For both MNIST and CIFAR-10, we observe that IBP shows decent performance at larger perturbation sizes, while other methods show limited improvement over IBP. This suggests that as perturbations increase in magnitude, stronger regularization is crucial for maintaining certifiability. In the context of corrupted datasets (MNIST-C and CIFAR-10-C), adversarial and certified training methods effectively enhance robustness against localized perturbations such as blur, noise, and pixelation. However, these methods remain less resilient to global transformations like brightness and contrast changes compared to standard training. This observation aligns with the intuition that adversarial and certified training primarily improve robustness in the immediate neighborhood of the original inputs, whereas global changes fall outside this region. More diverse data augmentation strategies may mitigate this problem, though this may also come at the cost of reduced certified adversarial accuracy.

When examining network-level properties such as neuron instability and network utilization, trends across datasets are less straightforward. In all cases, standard training results in the highest neuron instability, as expected due to the absence of regularization aimed at minimizing this effect. However, network utilization does not follow a consistent pattern. In some scenarios, certified training increases network utilization compared to adversarial training, indicating the learning of more complex patterns and relationships. Nevertheless, this trend is not universally observed, suggesting that the underlying dynamics of network utilization may be context-specific.

Overall, these findings highlight that while some performance trends persist across datasets, others are context-dependent, underscoring the need for context-specific analysis when evaluating the current stage of certified robustness methods.

\subsection{Comparing the OOD Generalization of Certified and Adversarial Training} \label{app:ood}

On corrupted datasets (MNIST-C, CIFAR-10-C), adversarial and certified training improve robustness against localized perturbations like blur, noise, and pixelation but struggle with global shifts like brightness and contrast changes. This aligns with the intuition that these methods enhance robustness mainly in the Euclidean neighborhood of the original inputs, whereas global changes fall outside this specification. Moreover, the stronger regularization induced by certified training when compared to adversarial training often exceeds what's needed for untargeted corruptions.

\subsection{Comparing our Loss Fragmentation Result with \citet{ShiWZYH21}} \label{app:loss_fragmentation}

While our approach and the results of \citet{ShiWZYH21} share common concepts and both target to quantify the difficulty of certification, we clarify that \citet{ShiWZYH21} analyze only IBP-bounded instability, which is an over-approximation of the real instability of neurons. In contrast, our analysis applies an estimate of the true number of unstable neurons. To illustrate this difference, we provide a comparison between the two variants in \cref{tab:unstable_neurons}. We observe that the gap between our lower bound estimate and IBP is larger for SOTA methods, which also reflects in the certification gap between IBP and MN-BAB for these models (\cref{tab:certification_incomplete}).

\begin{table}[h]
    \centering
    \caption{Unstable neuron estimates using our sampling method (lower bound) and IBP (upper bound) for networks trained on CIFAR-10 $\epsilon=2/255$.}
    \label{tab:unstable_neurons}
    \resizebox{0.4\linewidth}{!}{
        \begin{tabular}{lcc}
            \toprule
            \multirow{2.5}{*}{Method} & \multicolumn{2}{c}{Unstable Neurons [\%]}                     \\
            \cmidrule(rl){2-3}
                                      & lower bound                               & upper bound (IBP) \\
            \midrule
            IBP                       & 0.97                                      & 1.26              \\
            CROWN-IBP                 & 1.16                                      & 1.69              \\
            SABR                      & 1.37                                      & 2.82              \\
            TAPS                      & 1.00                                      & 1.68              \\
            STAPS                     & 1.32                                      & 2.64              \\
            MTL-IBP                   & 1.53                                      & 3.42              \\ \bottomrule
        \end{tabular}
    }
\end{table}

Both approaches aimed at quantifying neuron instability are hard to generalize to non-ReLU networks. However, this is of critical importance to the certification of ReLU networks, beyond measuring the difficulty of adversarial attacks which may also be indicated by other smoothness metrics. Concretely, branch-and-bound (BaB), the dominating strategy for complete certification of ReLU networks, directly branches the unstable neurons; thus, the number of unstable neurons provides a direct metric for the difficulty of certification. Since ReLU networks dominate certified training, we adopt the number of unstable neurons as the main metric.

\section{Experiment Details} \label{app:exp_setting}

\subsection{Dataset} \label{app:dataset}

We use the \mnist \citep{lecun2010mnist}, \cifar \citep{krizhevsky2009learning} and \TIN \citep{Ya2015tinyimagenet} datasets for our experiments. All are open-source and freely available with unspecified license.
The data preprocessing mostly follows \citet{palma2024expressive}. For MNIST, we do not apply any preprocessing. For CIFAR-10 and \TIN, we normalize with the dataset mean and standard deviation and augment with random horizontal flips. We apply random cropping to $32 \times 32$ after applying a $2$ pixel zero padding at every margin for \cifar, and random cropping to $64\times 64$ after applying a $4$ pixel zero padding at every margin for \TIN. We train on the corresponding train set and certify on the validation set, as adopted in the literature \citep{ShiWZYH21,MuellerEFV22,MaoM0V23,palma2024expressive}.

\subsection{Model Architectures} \label{app:model_architectures}

We follow \citet{ShiWZYH21,MuellerEFV22} and use a \cnns with Batch Norm for our main experiments. \cnns is a convolutional network with $7$ convolutional and linear layers. All but the last linear layer are followed by a Batch Norm and ReLU layer. This architecture is found to achieve uniformly better results across settings \citep{ShiWZYH21}, and thus is adopted by the literature \citep{ShiWZYH21,MuellerEFV22,MaoM0V23,palma2024expressive}. For \TIN, the stride of the last convolution is doubled to reduce the cost.

\subsection{Training Details} \label{app:training_details}

\paragraph{Initialization} Adversarial training methods are initialized by Kaiming uniform \citep{HeZRS15}, while certified training methods are initialized by \ibp initialization \citep{ShiWZYH21}.

\paragraph{Training Schedule} We mostly follow the training schedule of \citep{palma2024expressive}, but in some cases a shorter schedule to reduce cost. Specifically, the warmup phase is 20 epochs for \mnist $\epsilon=0.1$ and $\epsilon=0.3$, 80 epochs for \cifar $\epsilon=\frac{2}{255}$, 120 epochs for \cifar $\epsilon=\frac{8}{255}$ and 80 epochs for \TIN $\epsilon=\frac{1}{255}$. In addition, for \cifar and \TIN, we use standard training for 1 additional epoch at the beginning.
We apply the \ibp regularization proposed by \citep{ShiWZYH21}, with weight equals 0.5 on \mnist and \cifar, and 0.2 on \TIN, during the warmup phase. In total, we train 70 epochs for \mnist $\epsilon=0.1$ and $\epsilon=0.3$, 160 epochs for \cifar $\epsilon=\frac{2}{255}$, 240 epochs for \cifar $\epsilon=\frac{8}{255}$, and 160 epochs for \TIN $\epsilon=\frac{1}{255}$.

\paragraph{Optimization} We use Adam \citep{KingmaB14} with a learning rate of $0.0005$. The learning rate is decayed by a factor of $0.2$ at epoch 50 and 60 for \mnist $\epsilon=0.1$ and $\epsilon=0.3$, at epoch 120 and 140 for \cifar $\epsilon=\frac{2}{255}$, at epoch 200 and 220 for \cifar $\epsilon=\frac{8}{255}$, and at epoch 120 and 140 for \TIN $\epsilon=\frac{1}{255}$. We use a batch size of $256$ for \mnist, and $128$ for \cifar and \TIN. Gradients of each step are clipped to 10 in $L_2$ norm. No weight decay is applied and $L_1$ regularization only on weights of linear and convolution layers is used. Further, \citet{WuPreciseBN21} find that running statistics lag behind the population statistics and propose to use the population statistics for testing. We adopt this strategy in \ctbench, since it only needs to compute $\gL_{\nat}$ and is much cheaper than the computation of $\gL_{\rob}$.

\subsection{Tuning Scheme} \label{app:tuning_scheme}

We conduct a hyperparameter tuning for each method to ensure the best performance, and reduce the search space whenever appropriate based on human knowledge. The search space for each hyperparameter is as follows:
\begin{itemize}[leftmargin=*]
    \item \textit{$L_1$ regularization}: $\{1\times 10^{-6}, 2\times 10^{-6}, 5\times 10^{-6}, 1\times 10^{-5}, 2\times 10^{-5}, 5\times 10^{-5}\}$. We include $3\times 10^{-6}$ specifically for \cifar $\epsilon=\frac{2}{255}$, as this is the value reported by \citet{palma2024expressive}.
    \item \textit{$w_{\rob}$}: $\{0.7, 0.8, 0.9, 1.0\}$. Surprisingly, $w_{\rob}$ not equal to $1$ can improve both certified and natural accuracy by a large margin when $\epsilon$ is small.
    \item \textit{Train $\epsilon$}: we use 2x train $\epsilon$ for \mnist $\epsilon=0.1$, and tune within $\{$1x, 1.25x, 1.5x$\}$ specifically for \cifar $\epsilon=\frac{2}{255}$. For others, we use the test $\epsilon$ for training.
    \item \textit{$\epsilon$ shrink ratio for \sabr and \staps}: we mostly keep the value in the literature. When we observe large certifibility gap, we increase the shrink ratio by $0.1$ until the performance fails to increase consistently.
    \item \textit{Classifier size for \taps and \staps}: we keep the value in the literature for \taps, and include only 1 ReLU layer in the classifier for \staps universally.
    \item \textit{\taps gradient scale}: $\{1,2,3,4,6,8\}$.
    \item \textit{ReLU shrink ratio for \sabr and \staps}: we keep the value in the literature, thus shrinking the output box of each ReLU by multiplying 0.8 on \cifar $\epsilon=\frac{2}{255}$ and do not apply this in other settings.
    \item \textit{\ibp coefficient for \mtlibp}: $\{0.01, 0.02, 0.05\}$ for \mnist $\epsilon=0.1$, \cifar $\epsilon=\frac{2}{255}$ and \TIN $\epsilon=\frac{1}{255}$, and $\{0.4, 0.5, 0.6\}$ for \mnist $\epsilon=0.3$, \cifar $\epsilon=\frac{8}{255}$.
    \item \textit{Attack Strength}: we use 3 restarts everywhere for the attack. By default, we use 10 steps for \mnist $\epsilon=0.1$, 5 steps for \mnist $\epsilon=0.3$, 8 steps for \cifar $\epsilon=\frac{2}{255}$, 10 steps for \cifar $\epsilon=\frac{8}{255}$, and 1 step for \TIN $\epsilon=\frac{1}{255}$. However, we find \mtlibp benefits from using only 1 step everywhere, while more steps will hurt certified accuracy, thus we only use 1 step specifically for \mtlibp except \cifar $\epsilon=\frac{2}{255}$, consistent to \citet{palma2024expressive}. We further only use 2x attack $\epsilon$ for \mtlibp on \cifar $\epsilon=\frac{2}{255}$.
\end{itemize}

We report the best hyperparameter for each method respectively in \cref{tab:mnist0.1}, \cref{tab:mnist0.3}, \cref{tab:cifar2.255}, \cref{tab:cifar8.255}, and \cref{tab:tin1.255}.

\begin{table}
    \centering
    \caption{Best hyperparameter for \mnist $\epsilon=0.1$.}
    \label{tab:mnist0.1}
    \begin{adjustbox}{width=\linewidth,center}
        \begin{tabular}{lcccccccc}
            \toprule
                                    & \pgd              & \edac             & \ibp              & \crownibp         & \sabr             & \taps             & \staps            & \mtlibp           \\
            \midrule
            $L_1$ regularization    & $1\times 10^{-5}$ & $1\times 10^{-5}$ & $2\times 10^{-6}$ & $2\times 10^{-6}$ & $1\times 10^{-6}$ & $1\times 10^{-6}$ & $1\times 10^{-6}$ & $1\times 10^{-5}$ \\
            $w_{\rob}$              & 1.0               & 1.0               & 1.0               & 1.0               & 0.7               & 0.7               & 0.7               & 0.7               \\
            Train $\epsilon$        & 0.2               & 0.2               & 0.2               & 0.2               & 0.2               & 0.2               & 0.2               & 0.2               \\
            $\epsilon$ shrink ratio & /                 & /                 & /                 & /                 & 0.4               & /                 & 0.4               & /                 \\
            Classifier size         & /                 & /                 & /                 & /                 & /                 & 3                 & 1                 & /                 \\
            \taps gradient scale    & /                 & /                 & /                 & /                 & /                 & 4                 & 4                 & /                 \\
            ReLU shrink ratio       & /                 & /                 & /                 & /                 & /                 & /                 & /                 & /                 \\
            \ibp coefficient        & /                 & /                 & /                 & /                 & /                 & /                 & /                 & 0.02              \\
            \bottomrule
        \end{tabular}
    \end{adjustbox}
\end{table}

\begin{table}
    \centering
    \caption{Best hyperparameter for \mnist $\epsilon=0.3$.}
    \label{tab:mnist0.3}
    \begin{adjustbox}{width=\linewidth,center}
        \begin{tabular}{lcccccccc}
            \toprule
                                    & \pgd              & \edac             & \ibp              & \crownibp         & \sabr             & \taps             & \staps            & \mtlibp           \\
            \midrule
            $L_1$ regularization    & $5\times 10^{-6}$ & $5\times 10^{-6}$ & $1\times 10^{-6}$ & $1\times 10^{-6}$ & $2\times 10^{-6}$ & $2\times 10^{-6}$ & $2\times 10^{-6}$ & $1\times 10^{-6}$ \\
            $w_{\rob}$              & 1.0               & 1.0               & 1.0               & 1.0               & 1.0               & 1.0               & 1.0               & 1.0               \\
            Train $\epsilon$        & 0.3               & 0.3               & 0.3               & 0.3               & 0.3               & 0.3               & 0.3               & 0.3               \\
            $\epsilon$ shrink ratio & /                 & /                 & /                 & /                 & 0.8               & /                 & 0.8               & /                 \\
            Classifier size         & /                 & /                 & /                 & /                 & /                 & 1                 & 1                 & /                 \\
            \taps gradient scale    & /                 & /                 & /                 & /                 & /                 & 3                 & 1                 & /                 \\
            ReLU shrink ratio       & /                 & /                 & /                 & /                 & /                 & /                 & /                 & /                 \\
            \ibp coefficient        & /                 & /                 & /                 & /                 & /                 & /                 & /                 & 0.5               \\
            \bottomrule
        \end{tabular}
    \end{adjustbox}
\end{table}

\begin{table}
    \centering
    \caption{Best hyperparameter for \cifar  $\epsilon=2/255$.}
    \label{tab:cifar2.255}
    \begin{adjustbox}{width=\linewidth,center}
        \begin{tabular}{lcccccccc}
            \toprule
                                    & \pgd              & \edac              & \ibp              & \crownibp         & \sabr             & \taps             & \staps            & \mtlibp           \\
            \midrule
            $L_1$ regularization    & $2\times 10^{-5}$ & $5 \times 10^{-6}$ & $1\times 10^{-6}$ & $1\times 10^{-6}$ & $1\times 10^{-6}$ & $2\times 10^{-6}$ & $5\times 10^{-6}$ & $3\times 10^{-6}$ \\
            $w_{\rob}$              & 1.0               & 1.0                & 1.0               & 1.0               & 0.7               & 1.0               & 1.0               & 0.9               \\
            Train $\epsilon$        & $2/255$           & $2/255$            & $2/255$           & $2/255$           & $3/255$           & $2/255$           & $3/255$           & $2/255$           \\
            $\epsilon$ shrink ratio & /                 & /                  & /                 & /                 & 0.1               & /                 & 0.1               & /                 \\
            Classifier size         & /                 & /                  & /                 & /                 & /                 & 5                 & 1                 & /                 \\
            \taps gradient scale    & /                 & /                  & /                 & /                 & /                 & 5                 & 5                 & /                 \\
            ReLU shrink ratio       & /                 & /                  & /                 & /                 & 0.8               & /                 & 0.8               & /                 \\
            \ibp coefficient        & /                 & /                  & /                 & /                 & /                 & /                 & /                 & 0.01              \\
            \bottomrule
        \end{tabular}
    \end{adjustbox}
\end{table}

\begin{table}
    \centering
    \caption{Best hyperparameter for \cifar  $\epsilon=8/255$.}
    \label{tab:cifar8.255}
    \begin{adjustbox}{width=\linewidth,center}
        \begin{tabular}{lcccccccc}
            \toprule
                                    & \pgd              & \edac             & \ibp    & \crownibp & \sabr   & \taps   & \staps  & \mtlibp \\
            \midrule
            $L_1$ regularization    & $1\times 10^{-6}$ & $1\times 10^{-6}$ & $0$     & $0$       & $0$     & $0$     & $0$     & $0$     \\
            $w_{\rob}$              & 1.0               & 1.0               & 1.0     & 1.0       & 1.0     & 1.0     & 1.0     & 1.0     \\
            Train $\epsilon$        & $8/255$           & $8/255$           & $8/255$ & $8/255$   & $8/255$ & $8/255$ & $8/255$ & $8/255$ \\
            $\epsilon$ shrink ratio & /                 & /                 & /       & /         & 0.7     & /       & 0.9     & /       \\
            Classifier size         & /                 & /                 & /       & /         & /       & 1       & 1       & /       \\
            \taps gradient scale    & /                 & /                 & /       & /         & /       & 2       & 2       & /       \\
            ReLU shrink ratio       & /                 & /                 & /       & /         & /       & /       & /       & /       \\
            \ibp coefficient        & /                 & /                 & /       & /         & /       & /       & /       & 0.5     \\
            \bottomrule
        \end{tabular}
    \end{adjustbox}
\end{table}

\begin{table}
    \centering
    \caption{Best hyperparameter for \TIN $\epsilon=1/255$.}
    \label{tab:tin1.255}
    \begin{adjustbox}{width=\linewidth,center}
        \begin{tabular}{lcccccccc}
            \toprule
                                    & \pgd              & \edac             & \ibp              & \crownibp         & \sabr             & \taps             & \staps            & \mtlibp           \\
            \midrule
            $L_1$ regularization    & $5\times 10^{-5}$ & $1\times 10^{-5}$ & $1\times 10^{-5}$ & $1\times 10^{-5}$ & $1\times 10^{-5}$ & $1\times 10^{-5}$ & $1\times 10^{-5}$ & $5\times 10^{-5}$ \\
            $w_{\rob}$              & 1.0               & 1.0               & 1.0               & 1.0               & 1.0               & 1.0               & 1.0               & 0.7               \\
            Train $\epsilon$        & $1/255$           & $1/255$           & $1/255$           & $1/255$           & $1/255$           & $1/255$           & $1/255$           & $1/255$           \\
            $\epsilon$ shrink ratio & /                 & /                 & /                 & /                 & 0.4               & /                 & 0.6               & /                 \\
            Classifier size         & /                 & /                 & /                 & /                 & /                 & 1                 & 1                 & /                 \\
            \taps gradient scale    & /                 & /                 & /                 & /                 & /                 & 8                 & 4                 & /                 \\
            ReLU shrink ratio       & /                 & /                 & /                 & /                 & /                 & /                 & /                 & /                 \\
            \ibp coefficient        & /                 & /                 & /                 & /                 & /                 & /                 & /                 & 0.05              \\
            \bottomrule
        \end{tabular}
    \end{adjustbox}
\end{table}

\subsection{Certification Details} \label{app:certification_details}

We combine \ibp \citep{GowalIBP2018}, \crownibp\citep{ZhangCXGSLBH20}, and \mnbab\citep{FerrariMJV22} for certification running the most precise but also computationally costly \mnbab only on samples not certified by the other methods. The timout for each input is set to 1000 seconds.

\subsection{Computation} \label{app:computation}

We train and certify \mnist $\epsilon=0.1$, \mnist $\epsilon=0.3$ and \cifar $\epsilon=\frac{8}{255}$ models on a single NVIDIA GeForce RTX 2080 Ti with Intel(R) Xeon(R) Silver 4214R CPU @ 2.40GHz and 530GB RAM. We train and certify \cifar $\epsilon=\frac{2}{255}$ and \TIN $\epsilon=\frac{1}{255}$ models on a single NVIDIA L4 with Intel(R) Xeon(R) CPU @ 2.20GHz CPU and 377 GB RAM. The training and certification time for each method is reported in \cref{tb:time}. We provide a detailed complexity analysis for each training method in \cref{tab:complexity}.

\begin{table}[h]
    \centering
    \captionof{table}{Detailed breakdown of training costs for each Certified Training method.}
    \label{tab:complexity}
    \resizebox{0.95\linewidth}{!}{
        \begin{tabular}{lll}
            \toprule
            Method         & Training cost per batch     & Details                                                             \\
            \midrule
            Standard       & T                           & Forward + Backward                                                  \\
            PGD / EDAC     & (M+1)T                      & M attack steps + Standard loss computation                          \\
            IBP            & 2T                          & Lower and Upper Bounds propagation                                  \\
            CROWN-IBP + LF & 4T                          & IBP pass + back-substitution of IBP bounds                          \\
            SABR           & (M+2)T                      & IBP + PGD                                                           \\
            MTL-IBP        & (M+2)T                      & IBP + PGD                                                           \\
            TAPS           & 2t + K*(M+1)*(T-t)          & IBP for first network split and PGD for second split for each class \\
            STAPS          & 2t + K*(M+1)*(T-t) + (M+1)T & TAPS + PGD                                                          \\
            \midrule
            \multicolumn{3}{c}{Legend}                                                                                         \\
            \midrule
                           & T                           & Time cost for Standard Training (Includes Forward + Backward Pass)  \\
                           & M                           & Number of adversarial attack steps (including repeats)              \\
                           & K                           & Number of classes                                                   \\
                           & t                           & Time cost for Standard Training in the first network split (TAPS)   \\
            \bottomrule
        \end{tabular}
    }
\end{table}

\begin{table}
    \centering
    \caption{Training and certification time for each method on different datasets and $\epsilon$.} \label{tb:time}
    \begin{adjustbox}{width=.75\linewidth,center}
        \begin{tabular}{ccccc}
            \toprule
            Dataset                & $\epsilon$                     & Method    & Train Time (seconds) & Certification Time (seconds) \\
            \midrule
            \multirow{16}*{\mnist} & \multirow{8}*{$0.1$}           & \pgd      & $1.5 \times 10^4$    & /                            \\
                                   &                                & \edac     & $3.1 \times 10^4$    & /                            \\
                                   &                                & \ibp      & $2.1 \times 10^3$    & $2.5 \times 10^3$            \\
                                   &                                & \crownibp & $5.6 \times 10^3$    & $1.8 \times 10^3$            \\
                                   &                                & \sabr     & $1.8 \times 10^4$    & $6.0 \times 10^3$            \\
                                   &                                & \taps     & $3.8 \times 10^4$    & $6.0 \times 10^3$            \\
                                   &                                & \staps    & $2.5 \times 10^4$    & $6.9 \times 10^3$            \\
                                   &                                & \mtlibp   & $6.8 \times 10^3$    & $6.8 \times 10^3$            \\
            \cmidrule{2-5}
                                   & \multirow{8}*{$0.3$}           & \pgd      & $1.1 \times 10^4$    & /                            \\
                                   &                                & \edac     & $2.2 \times 10^4$    & /                            \\
                                   &                                & \ibp      & $2.6 \times 10^3$    & $3.2 \times 10^4$            \\
                                   &                                & \crownibp & $5.4 \times 10^3$    & $2.6 \times 10^4$            \\
                                   &                                & \sabr     & $9.7 \times 10^3$    & $5.2 \times 10^4$            \\
                                   &                                & \taps     & $7.1 \times 10^3$    & $4.7 \times 10^4$            \\
                                   &                                & \staps    & $1.4 \times 10^4$    & $5.1 \times 10^4$            \\
                                   &                                & \mtlibp   & $5.5 \times 10^3$    & $4.4 \times 10^4$            \\
            \midrule
            \multirow{16}*{\cifar} & \multirow{8}*{$\frac{2}{255}$} & \pgd      & $2.8 \times 10^4$    & /                            \\
                                   &                                & \edac     & $1.3 \times 10^5$    & /                            \\
                                   &                                & \ibp      & $1.2 \times 10^4$    & $1.3 \times 10^5$            \\
                                   &                                & \crownibp & $2.7 \times 10^4$    & $1.9 \times 10^5$            \\
                                   &                                & \sabr     & $2.4 \times 10^4$    & $1.6 \times 10^5$            \\
                                   &                                & \taps     & $1.1 \times 10^5$    & $1.1 \times 10^5$            \\
                                   &                                & \staps    & $4.5 \times 10^4$    & $3.0 \times 10^5$            \\
                                   &                                & \mtlibp   & $3.6 \times 10^4$    & $2.7 \times 10^5$            \\
            \cmidrule{2-5}
                                   & \multirow{8}*{$\frac{8}{255}$} & \pgd      & $6.4 \times 10^4$   & /                            \\
                                   &                                & \edac     & $1.3 \times 10^5$   & /                            \\
                                   &                                & \ibp      & $1.1 \times 10^4$   & $1.9 \times 10^4$            \\
                                   &                                & \crownibp & $2.1 \times 10^4$   & $2.0 \times 10^4$            \\
                                   &                                & \sabr     & $4.1 \times 10^4$   & $6.5 \times 10^4$            \\
                                   &                                & \taps     & $3.3 \times 10^4$   & $4.0 \times 10^4$            \\
                                   &                                & \staps    & $9.9 \times 10^4$   & $4.2 \times 10^4$            \\
                                   &                                & \mtlibp   & $2.2 \times 10^4$   & $5.6 \times 10^4$            \\
            \midrule
            \multirow{8}*{\TIN}    & \multirow{8}*{$\frac{1}{255}$} & \pgd      & $1.0 \times 10^5$    & /                            \\
                                   &                                & \edac     & $2.0 \times 10^5$    & /                            \\
                                   &                                & \ibp      & $6.7 \times 10^4$    & $4.9 \times 10^3$            \\
                                   &                                & \crownibp & $2.0 \times 10^5$    & $1.3 \times 10^4$            \\
                                   &                                & \sabr     & $1.1 \times 10^5$    & $1.8 \times 10^4$            \\
                                   &                                & \taps     & $2.8 \times 10^5$    & $1.5 \times 10^4$            \\
                                   &                                & \staps    & $3.3 \times 10^5$    & $2.6 \times 10^4$            \\
                                   &                                & \mtlibp   & $1.5 \times 10^5$    & $5.1 \times 10^3$            \\
            \bottomrule
        \end{tabular}
    \end{adjustbox}
\end{table}

\clearpage
\section{Additional Results} \label{app:additional_results}

\subsection{Result Significance} \label{app:training_stability}

In \cref{tab:randomness_mnist_01,tab:randomness_mnist_03,tab:randomness_cifar_2,tab:randomness_cifar_8,tab:randomness_tin}, we present the natural and certified accuracy across certified training algorithms, datasets and perturbation levels. We report the average and standard deviation across 3 random seeds for each method. The results show that our improvements over previously reported values in the literature are significant, in most cases the statistical difference being larger than $3\sigma$.

\begin{table}[h]
    \caption{Comparison of Natural and Certified Accuracy between \ctbench and previous literature results on \mnist $\epsilon=0.1$. We report average and standard deviation across 3 random seeds for each method.} \label{tab:randomness_mnist_01}
    \begin{adjustbox}{width=.5\linewidth,center}
        \begin{tabular}{@{}clcc@{}}
            \toprule
            {Method}                   & Source     & Nat. [\%]        & Cert. [\%]       \\ \midrule
            \multirow{2}{*}{IBP}       & Literature & 98.84            & 97.95            \\
                                       & This work  & 98.86 $\pm$ 0.06 & 98.25 $\pm$ 0.03 \\ \midrule
            \multirow{2}{*}{CROWN-IBP} & Literature & 98.83            & 97.76            \\
                                       & This work  & 98.93 $\pm$ 0.01 & 98.17 $\pm$ 0.05 \\ \midrule
            \multirow{2}{*}{SABR}      & Literature & 99.23            & 98.22            \\
                                       & This work  & 99.15 $\pm$ 0.08 & 98.42 $\pm$ 0.03 \\ \midrule
            \multirow{2}{*}{TAPS}      & Literature & 99.19            & 98.39            \\
                                       & This work  & 99.20 $\pm$ 0.05 & 98.5 $\pm$ 0.04  \\ \midrule
            \multirow{2}{*}{STAPS}     & Literature & 99.15            & 98.37            \\
                                       & This work  & 99.15 $\pm$ 0.04 & 98.38 $\pm$ 0.10 \\ \midrule
            \multirow{2}{*}{MTL-IBP}   & Literature & 99.25            & 98.38            \\
                                       & This work  & 99.16 $\pm$ 0.03 & 98.31 $\pm$ 0.06 \\ \bottomrule
        \end{tabular}
    \end{adjustbox}
\end{table}

\begin{table}[h]
    \caption{Comparison of Natural and Certified Accuracy between \ctbench and previous literature results on \mnist $\epsilon=0.3$. We report average and standard deviation across 3 random seeds for each method.} \label{tab:randomness_mnist_03}
    \begin{adjustbox}{width=.5\linewidth,center}
        \begin{tabular}{@{}clcc@{}}
            \toprule
            Method                     & Source     & Nat. {[}\%{]}    & Cert. {[}\%{]}   \\ \midrule
            \multirow{2}{*}{IBP}       & Literature & 97.67            & 93.10            \\
                                       & This work  & 98.55 $\pm$ 0.02 & 93.82 $\pm$ 0.10 \\ \midrule
            \multirow{2}{*}{CROWN-IBP} & Literature & 98.18            & 92.98            \\
                                       & This work  & 98.46 $\pm$ 0.03 & 93.84 $\pm$ 0.12 \\ \midrule
            \multirow{2}{*}{SABR}      & Literature & 98.75            & 93.40            \\
                                       & This work  & 98.69 $\pm$ 0.03 & 93.64 $\pm$ 0.06 \\ \midrule
            \multirow{2}{*}{TAPS}      & Literature & 97.94            & 93.62            \\
                                       & This work  & 98.58 $\pm$ 0.03 & 93.90 $\pm$ 0.11 \\ \midrule
            \multirow{2}{*}{STAPS}     & Literature & 98.53            & 93.51            \\
                                       & This work  & 98.69 $\pm$ 0.06 & 93.60 $\pm$ 0.05 \\ \midrule
            \multirow{2}{*}{MTL-IBP}   & Literature & 98.80            & 93.62            \\
                                       & This work  & 98.75 $\pm$ 0.02 & 93.82 $\pm$ 0.21 \\ \bottomrule
        \end{tabular}
    \end{adjustbox}
\end{table}

\begin{table}[h]
    \caption{Comparison of Natural and Certified Accuracy between \ctbench and previous literature results on \cifar $\epsilon=2/255$. We report average and standard deviation across 3 random seeds for each method.} \label{tab:randomness_cifar_2}
    \begin{adjustbox}{width=.5\linewidth,center}
        \begin{tabular}{@{}clcc@{}}
            \toprule
            Method                     & Source     & Nat. {[}\%{]}    & Cert. {[}\%{]}   \\ \midrule
            \multirow{2}{*}{IBP}       & Literature & 66.84            & 52.85            \\
                                       & This work  & 66.85 $\pm$ 0.72 & 55.32 $\pm$ 0.68 \\ \midrule
            \multirow{2}{*}{CROWN-IBP} & Literature & 71.52            & 53.97            \\
                                       & This work  & 67.56 $\pm$ 0.04 & 56.69 $\pm$ 0.58 \\ \midrule
            \multirow{2}{*}{SABR}      & Literature & 79.24            & 62.84            \\
                                       & This work  & 77.82 $\pm$ 0.28 & 63.62 $\pm$ 0.22 \\ \midrule
            \multirow{2}{*}{TAPS}      & Literature & 75.09            & 61.56            \\
                                       & This work  & 74.76 $\pm$ 0.34 & 61.37 $\pm$ 0.09 \\ \midrule
            \multirow{2}{*}{STAPS}     & Literature & 79.76            & 62.98            \\
                                       & This work  & 76.88 $\pm$ 0.15 & 63.96 $\pm$ 0.27 \\ \midrule
            \multirow{2}{*}{MTL-IBP}   & Literature & 80.11            & 63.24            \\
                                       & This work  & 78.91 $\pm$ 0.16 & 64.00 $\pm$ 0.37 \\ \bottomrule
        \end{tabular}
    \end{adjustbox}
\end{table}

\begin{table}[h]
    \caption{Comparison of Natural and Certified Accuracy between \ctbench and previous literature results on \cifar $\epsilon=8/255$. We report average and standard deviation across 3 random seeds for each method.} \label{tab:randomness_cifar_8}
    \begin{adjustbox}{width=.5\linewidth,center}
        \begin{tabular}{@{}clcc@{}}
            \toprule
            Method                     & Source     & Nat. {[}\%{]}    & Cert. {[}\%{]}   \\ \midrule
            \multirow{2}{*}{IBP}       & Literature & 48.94            & 34.97            \\
                                       & This work  & 48.74 $\pm$ 0.23 & 34.99 $\pm$ 0.28 \\ \midrule
            \multirow{2}{*}{CROWN-IBP} & Literature & 46.29            & 33.38            \\
                                       & This work  & 48.24 $\pm$ 0.09 & 32.49 $\pm$ 0.18 \\ \midrule
            \multirow{2}{*}{SABR}      & Literature & 52.38            & 35.13            \\
                                       & This work  & 52.51 $\pm$ 0.38 & 34.97 $\pm$ 0.62 \\ \midrule
            \multirow{2}{*}{TAPS}      & Literature & 49.76            & 35.10            \\
                                       & This work  & 49.82 $\pm$ 0.28 & 34.89 $\pm$ 0.40 \\ \midrule
            \multirow{2}{*}{STAPS}     & Literature & 52.82            & 34.65            \\
                                       & This work  & 51.46 $\pm$ 0.25 & 35.32 $\pm$ 0.25 \\ \midrule
            \multirow{2}{*}{MTL-IBP}   & Literature & 53.35            & 35.44            \\
                                       & This work  & 53.72 $\pm$ 0.49 & 35.23 $\pm$ 0.18 \\ \bottomrule
        \end{tabular}
    \end{adjustbox}
\end{table}

\clearpage
\begin{table}[h]
    \caption{Comparison of Natural and Certified Accuracy between \ctbench and previous literature results on \TIN $\epsilon=1/255$. We report average and standard deviation across 3 random seeds for each method.} \label{tab:randomness_tin}
    \begin{adjustbox}{width=.5\linewidth,center}
        \begin{tabular}{@{}clcc@{}}
            \toprule
            Method                     & Source     & Nat. {[}\%{]}    & Cert. {[}\%{]}   \\ \midrule
            \multirow{2}{*}{IBP}       & Literature & 25.92            & 17.87            \\
                                       & This work  & 26.4 $\pm$ 0.45  & 19.87 $\pm$ 0.19 \\ \midrule
            \multirow{2}{*}{CROWN-IBP} & Literature & 25.62            & 17.93            \\
                                       & This work  & 28.16 $\pm$ 0.27 & 21.69 $\pm$ 0.42 \\ \midrule
            \multirow{2}{*}{SABR}      & Literature & 28.85            & 20.46            \\
                                       & This work  & 30.96 $\pm$ 0.41 & 21.14 $\pm$ 0.2  \\ \midrule
            \multirow{2}{*}{TAPS}      & Literature & 28.34            & 20.82            \\
                                       & This work  & 28.59 $\pm$ 0.09 & 21.54 $\pm$ 0.22 \\ \midrule
            \multirow{2}{*}{STAPS}     & Literature & 28.98            & 22.16            \\
                                       & This work  & 30.25 $\pm$ 0.33 & 22.03 $\pm$ 0.25 \\ \midrule
            \multirow{2}{*}{MTL-IBP}   & Literature & 37.56            & 26.09            \\
                                       & This work  & 35.97 $\pm$ 0.17 & 27.49 $\pm$ 0.21 \\ \bottomrule
        \end{tabular}
    \end{adjustbox}
\end{table}

\subsection{Architecture Generalization} \label{app:expressive_comparison}

In \cref{tab:expressive_comparison} we present the natural and certified accuracy on \cnnf, which is a five-layer CNN smaller than \cnns. We observe that the improvements are consistent across different settings even with \cnnf, showing that the improvements are not specific to a certain architecture.

\begin{table}[h]
    \caption{Comparison on \cnnf between \ctbench and the implementation of \citet{palma2024expressive}.}\label{tab:expressive_comparison}
    \begin{adjustbox}{width=.5\linewidth,center}
        \begin{tabular}{@{}clcc@{}}
            \toprule
            Method                   & Code and hyperparameters   & Nat. {[}\%{]} & Cert. {[}\%{]} \\ \midrule
            \multirow{2}{*}{IBP}     & \ctbench                   & 98.19         & 92.88          \\
                                     & \cite{palma2024expressive} & 93.16         & 81.81          \\ \midrule
            \multirow{2}{*}{SABR}    & \ctbench                   & 98.41         & 92.62          \\
                                     & \cite{palma2024expressive} & 97.33         & 90.87          \\ \midrule
            \multirow{2}{*}{MTL-IBP} & \ctbench                   & 98.41         & 92.49          \\
                                     & \cite{palma2024expressive} & 98.39         & 91.45          \\ \bottomrule
        \end{tabular}
    \end{adjustbox}
\end{table}

\clearpage
\subsection{Ablation on Certification Algorithms \& Additional Analysis on Shared Mistakes}\label{app:correlation}

In \cref{tb:common_mistake_cifar_certification} we present the correlation between the certification capabilities of two SOTA verifiers (\mnbab \citep{FerrariMJV22} and OVAL \citep{PalmaIBPR22}). We observe that there is a very high correlation between the two verifiers, which is expected since both are based on the same underlying principles. This shows that the certification algorithms have reached a certain level of maturity and are converging to similar results. Combining the verified sets of the two verifiers, we get a marginal improvement in certified accuracy at the cost of a much larger certification time. Therefore, we only use \mnbab for certification in our main experiments.

\begin{table}[h]
    \centering
    \caption{Observed count of common mistakes of certification algorithms (\mnbab \citep{FerrariMJV22} and OVAL \citep{PalmaIBPR22}) on \cifar against their expected values assuming independence across certification mistakes.} \label{tb:common_mistake_cifar_certification}
    \begin{adjustbox}{width=.55\linewidth,center}
        \begin{tabular}{ccccc}
            \toprule
                                            &      & neither certify & one certifies & both certify \\
            \cmidrule{3-5}
            \multirow{2}*{$\epsilon=2/255$} & obs. & 3549            & 15            & 6436         \\
                                            & exp. & 1264            & 4585          & 4151         \\
            \midrule
            \multirow{2}*{$\epsilon=8/255$} & obs. & 6454            & 9             & 3537         \\
                                            & exp. & 4171            & 4575          & 1254         \\

            \bottomrule
        \end{tabular}
    \end{adjustbox}
    \vspace{-3mm}
\end{table}

In \cref{tb:common_mistake_cifar} we present the observed count of common mistakes that different certified training models make on \cifar against their expected values assuming independence across model mistakes. We observe that the observed count is significantly higher than the expected count, indicating that the models are highly correlated in their mistakes.

\begin{table}[h]
    \centering
    \caption{Observed count of common mistakes on \cifar against their expected values assuming independence across model mistakes.} \label{tb:common_mistake_cifar}
    \begin{adjustbox}{width=.6\linewidth,center}
        \begin{tabular}{ccccccccc}
            \toprule
                                                    &      & \multicolumn{7}{c}{\# models succeeded}                                           \\
                                                    &      & 0                                       & 1    & 2    & 3    & 4    & 5    & 6    \\
            \cmidrule{3-9}
            \multirow{2}*{$\epsilon=\frac{2}{255}$} & obs. & 2350                                    & 653  & 520  & 564  & 708  & 894  & 4311 \\
                                                    & exp. & 35                                      & 330  & 1296 & 2704 & 3163 & 1965 & 507  \\
            \midrule
            \multirow{2}*{$\epsilon=\frac{8}{255}$} & obs. & 5206                                    & 679  & 487  & 388  & 387  & 585  & 2268 \\
                                                    & exp. & 766                                     & 2457 & 3283 & 2339 & 937  & 200  & 18   \\
            \bottomrule
        \end{tabular}
    \end{adjustbox}
\end{table}

Furthermore, \cref{tab:common_mistake_cnn5} examines shared mistakes between CNN5 and CNN7, revealing common patterns across architectures.

\begin{table}[h]
    \centering
    \caption{Additional results on CNN5 and CNN7 shared mistakes for MNIST 0.3}
    \label{tab:common_mistake_cnn5}
    \resizebox{0.5\linewidth}{!}{
        \begin{tabular}{@{}lcc@{}}
            \toprule
            \multicolumn{1}{c}{\multirow{2}{*}{Models}} & \multicolumn{2}{c}{Number of not certified samples}                           \\
            \cmidrule(r){2-3}
            \multicolumn{1}{c}{}                        & Observed                                            & Expected if independent \\
            \midrule
            CNN5 IBP                                    & 771                                                 & /                       \\
            CNN5 SABR                                   & 793                                                 & /                       \\
            CNN5 MTL-IBP                                & 746                                                 & /                       \\
            CNN7 IBP                                    & 620                                                 & /                       \\
            CNN7 SABR                                   & 632                                                 & /                       \\
            CNN7 MTL-IBP                                & 610                                                 & /                       \\
            \midrule
            CNN5, CNN7 IBP                              & 526                                                 & 48                      \\
            CNN5, CNN7 SABR                             & 541                                                 & 50                      \\
            CNN5, CNN7 MTL-IBP                          & 516                                                 & 46                      \\
            All 3 CNN5 networks                         & 593                                                 & 5                       \\
            \bottomrule
        \end{tabular}
    }
\end{table}

\subsection{Deferred Results on \cifar} \label{app:additional_cifar}

\begin{figure}[h]
    \centering
    \vspace{-2mm}
    \includegraphics[width=.68\linewidth]{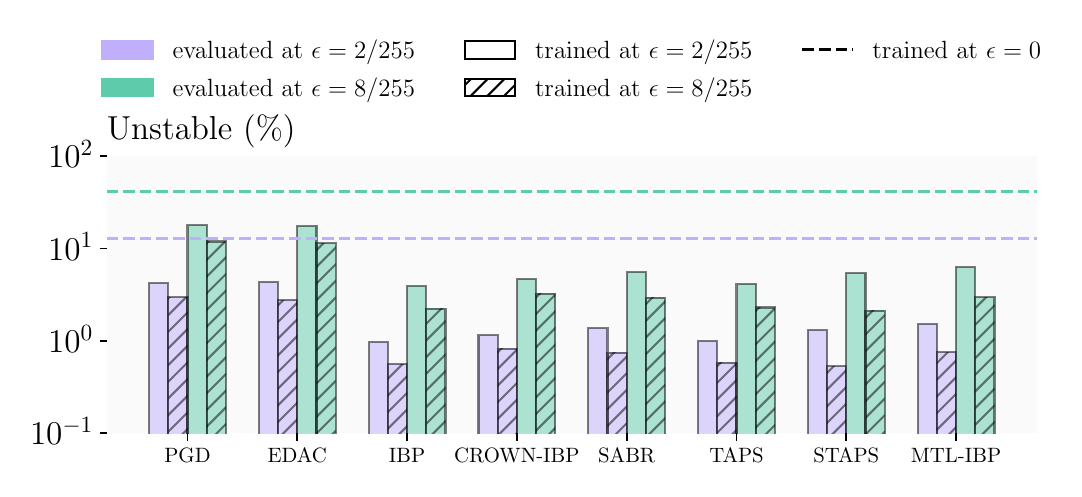}
    \caption{Ratio of unstable neurons for models trained on \cifar with different methods and $\epsilon$.} \label{fig:fragmentation_cifar}
    \vspace{-2mm}
\end{figure}

\begin{figure}[h]
    \centering
    \includegraphics[width=.7\linewidth]{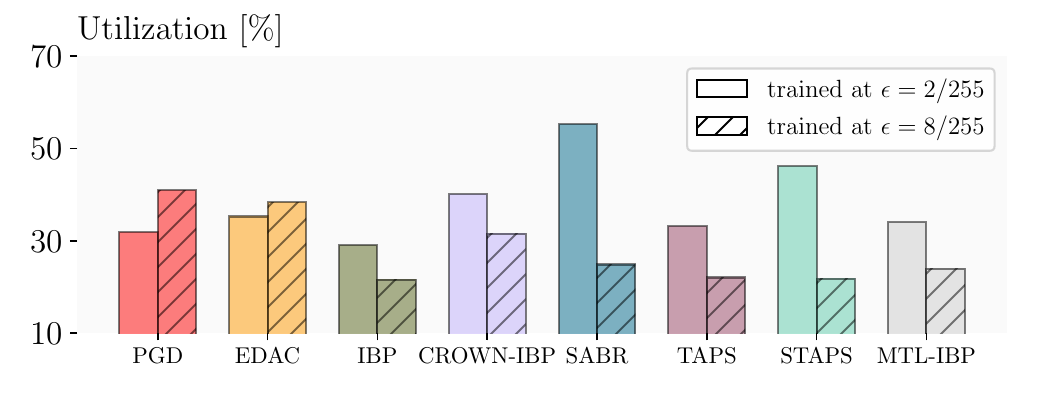}
    \caption{Model utilization for models trained on \cifar with different methods and $\epsilon$. We note that standard training has 35.79\% utilization.} \label{fig:utilization_cifar}
\end{figure}

In \cref{fig:fragmentation_cifar,fig:utilization_cifar}, we present additional analyses on the neuron statistics for different models trained on \cifar. We analyze the amount of unstable neurons and the model utilization for each model.

\begin{figure}[h]
    \centering
    \vspace{-3mm}
    \adjustbox{minipage=[c]{0.7\linewidth}}{

        \includegraphics[width=\linewidth]{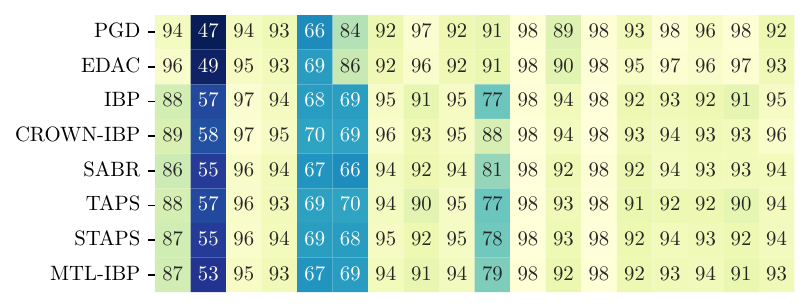}

        \includegraphics[width=\linewidth]{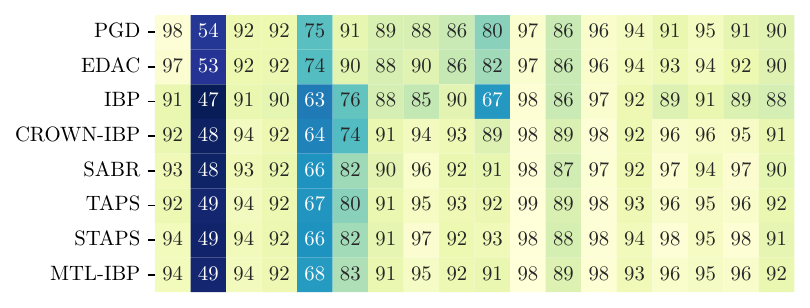}

        \hfill
        \includegraphics[width=.948\linewidth]{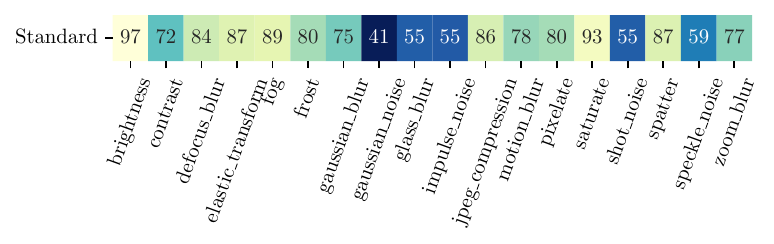}

    }
    \caption{Out-of-distribution generalization evaluated on \cifar-C for models trained on \cifar at $\epsilon=8/255$ (top), $\epsilon=2/255$ (middle) and standard training (bottom).} \label{fig:ood_cifar}
    \vspace{-8mm}
\end{figure}

In \cref{fig:ood_cifar}, we present the out-of-distribution generalization evaluated on \cifar-C for models trained on \cifar at $\epsilon=8/255$, $\epsilon=2/255$ and standard training. We observe that the models trained with certified training methods have better out-of-distribution generalization compared to standard training.

\subsection{Results with Incomplete Certification Algorithms} \label{app:certification_incomplete}

In \cref{tab:certification_incomplete}, we present the results of the certification with more efficient, but incomplete methods (IBP and CROWN-IBP). We observe that the incomplete methods have a significant impact on the certified accuracy, and the results are consistent with the previous findings in the literature.

\begin{table}
    \centering
    \caption{A comparison of incomplete certification (IBP and CROWN-IBP) against complete certification (MN-BAB) on CIFAR-10 $\epsilon=2/255$.}
    \label{tab:certification_incomplete}
    \resizebox{0.5\linewidth}{!}{
        \begin{tabular}{@{}lccccc@{}}
            \toprule
            \multirow{2.5}{*}{Train Method} & \multirow{2.5}{*}{Nat [\%]} & \multicolumn{3}{c}{Cert [\%]} & \multirow{2.5}{*}{Adv [\%]}                                   \\
            \cmidrule(lr){3-5}
                                            &                             & IBP                           & CROWN-IBP                   & MN-BAB         &                \\
            \midrule
            IBP                             & 67.49                       & \textbf{54.22}                & 54.57                       & 55.99          & 56.10          \\
            CROWN-IBP                       & 67.60                       & 49.92                         & \textbf{54.86}              & 57.11          & 57.28          \\
            SABR                            & 77.86                       & 12.12                         & 44.79                       & 63.61          & 65.56          \\
            TAPS                            & 74.44                       & 28.22                         & 50.14                       & 61.27          & 62.62          \\
            STAPS                           & 77.05                       & 0.72                          & 30.92                       & 64.21          & 66.09          \\
            MTL-IBP                         & \textbf{78.82}              & 0.62                          & 23.31                       & \textbf{64.41} & \textbf{67.69} \\ \bottomrule
        \end{tabular}
    }
\end{table}

\clearpage

\end{document}